\documentclass[twoside,11pt]{article}

\usepackage{arxiv}
\usepackage{amsmath, amsfonts}
\usepackage{natbib}
\usepackage{microtype}

\usepackage[capitalise,nameinlink,noabbrev]{cleveref}
\usepackage{algorithm2e}
\usepackage{color}
\usepackage{booktabs}
\usepackage{thmtools}
\usepackage{thm-restate}

\RequirePackage{algorithm}
\RequirePackage{algorithmic}

\DeclareMathOperator*{\argmax}{arg\,max}

\newcommand{\bm}[1]{\mathbf{#1}}
\newtheorem{assumption}{Assumption}

\ShortHeadings{Learning Causal State Representations of Partially Observable Environments}{Zhang et al.}
\firstpageno{1}

\begin{document}

\title{Learning Causal State Representations of Partially Observable Environments}

\author{
    \name Amy Zhang \email amy.x.zhang@mail.mcgill.ca \\
    \addr McGill University  \\
    Facebook AI Research
    \AND 
    \name Zachary C. Lipton  \\
    \addr Carnegie Mellon University
    \AND 
    \name Luis Pineda \\
    \addr Facebook AI Research
    \AND 
    \name Kamyar Azizzadenesheli \\
    \addr Purdue University
    \AND
    \name Anima Anandkumar \\
    \addr California Institute of Technology 
    \AND 
    \name Laurent Itti \\
    \addr University of Southern California
    \AND 
    \name Joelle Pineau  \\
    \addr McGill  University \\
    Facebook AI Research
    \AND
    \name Tommaso Furlanello \email tf@hk3lab.ai  \\
    \addr HK3 Lab
}

\editor{}

\maketitle

\begin{abstract}
Intelligent agents can cope with sensory-rich environments
by learning task-agnostic state abstractions. In this paper, we propose an algorithm to approximate \textit{causal states}, which are the coarsest partition of the joint history of actions and observations in partially-observable Markov decision processes (POMDP). Our method learns approximate causal state representations from RNNs trained to predict subsequent observations given the history. We demonstrate that these learned state representations are useful for learning policies efficiently in reinforcement learning problems with rich observation spaces. We connect causal states with causal feature sets from the causal inference  literature, and also provide theoretical guarantees on the optimality of the continuous version of this causal state representation under Lipschitz assumptions by proving equivalence to bisimulation, a relation between behaviorally equivalent systems. This allows for lower bounds on the optimal value function of the learned representation, which is tight given certain assumptions. Finally, we empirically evaluate causal state representations using multiple partially observable tasks and compare with prior methods. 
\end{abstract}


\section{Introduction}
Decision-making and control often require 
that an agent interact with a partially-observed environment
to learn an approximation of the true states and the underlying dynamics. 
To enable efficient planning, we must construct 
latent representations of sequences of (action, observation) trajectories. 
At present, there are three general approaches for learning these latent representations. All three methods infer structure in the sequence of actions and observations to learn an optimal policy.  The first is based on the class POMDP formalism, where dynamics of a domain are captured by state-to-state transition probability distributions and state-conditional observation density functions \citep{aastrom1965optimal,cassandra1994acting,veness2011aixi,roy2005compressedsensing,azizzadenesheli2016reinforcement}.  The second is framed in terms of predictive state representations (PSRs), 
where a systems dynamic matrix is used to model the trajectories of observations
\citep{littman2002predictive,singh2004predictive}, and statistical or spectral methods are used to find a low-dimensional approximation. The third uses Recurrent Neural Networks (RNNs) \citep{schmidhuber1990rnnpomdp,Schmidhuber90anon-line}, to learn policies for partially observable environments via gradient methods~\citep{hausknecht2015deep}. Though several approaches have been explored, we still lack methods that can handle high-dimensional observations spaces, while providing theoretical guarantees for the representation they find. We also are short of understanding the relation between several of these methods and the causal learning literature, which has also begun to address the problem of recovering representations for planning and decision-making. In sum, we propose a fourth approach for learning latent representations in partial observability settings with causal states, and provide theoretical guarantees from both computational mechanics and causal inference literature.

\subsection*{Summary of Contributions}
The three main contributions of our work are 
(1) a new connection between PSRs and bisimulation via causal states;
(2) a novel gradient-based algorithm for training a causal state representation;
and (3) a bound on the optimal value function of this causal state representation. 

We propose a principled approach for learning state representations
that generalizes PSRs to RNN-based methods. We exploit the idea of \textit{causal states}, 
which were introduced in the computational mechanics literature
and are defined as the coarsest partition of histories into classes
that are maximally predictive of the future \citep{crutchfield1989inferring}.
Causal states constitute a discrete causal graph, 
and can be described by a hidden Markov model with unifilar dynamics, i.e. there is no conditional entropy on the next state once the next observation is known. 

Causal states also have a clear connection to \textit{bisimulation}, 
a binary relation between behaviorally equivalent systems~\citep{ferns2004bisimulation}. 
Bisimulation relations are a type of state abstraction which offer a mathematically precise definition of what it means for two environments to `share the same structure'~\citep{larsen1989bisim,Givan2003EquivalenceNA}. We say that two states are bisimilar if they share the same expected reward and equivalent distributions over next bisimilar states. 
We show that causal states are equivalent to 
the coarsest bisimilar partition of the environment. Further, one can define bisimulation metrics, which induce an analogous notion of behavioural distance between states in an MDP.
This connection allows us to obtain a lower bound on the optimal value function 
of the learned causal state representation which relies on this metric.  This bound is derived from the distance between the learned states and the original POMDP. 
In summary, causal states form an MDP that is behaviorally equivalent to the observed POMDP and guaranteed to be the coarsest partitioned MDP in the family of behaviorally equivalent MDPs. 

We provide a gradient-based algorithm for learning causal state representations, and evaluate our algorithm for approximate causal states reconstruction on multiple GridWorld navigation tasks with partially observable states. 
We further evaluate approximate causal states on environments with continuous latent states, 
a modification of the original VizDoom environment used in \citep{koul2018learning}, and flickering Atari tasks~\citep{bellemare13arcade} to show that our algorithm is also robust to this setting.

\section{Background}

In this section we present notation and a general overview of concepts used throughout this paper, starting with state representations for decision processes, causal states as defined in computational mechanics, causal inference via invariant prediction, and state abstractions and bisimulation. 
\subsection{State Representations for Decision Processes}
\label{sec:decision_proc}
Consider the nonlinear stochastic process
emerging from the interaction between an agent's policy that 
chooses discrete actions $A_{t}$, taking values $a_{t}$ 
from the alphabet $\mathbf{A}$, and an environmental response $O_{t+1}$, 
taking values 
from the alphabet $\mathbf{O}$. 
Let $Y_t:=(O_t,A_{t-1})$ be the joint observation-action variable 
with realizations $y_t=(o_t,a_{t-1})$, 
where $y_t\in \mathbf{O}\times\mathbf{A}$,\footnote{For 
a fixed random variable ${Y_t}$, 
we indicate its inclusive past 
with $Y_{\leq t}=...Y_{t-3},Y_{t-2}, Y_{t-1} Y_{t}$ 
and its non-inclusive future with  $Y_{> t}=Y_{t+1}, Y_{t+2}...$,
dropping the subscript $t$ from the notation for convenience 
when the context is clear.}.
The future dynamics $\mathbb{P}(Y_{> t})=\mathbb{P}(O_{> t},A_{\geq t})$ 
depend jointly on the stationary \textit{policy}
$\mathbb{P}(A_{\geq t}|Y_{\leq t})$ 
that maps the joint histories $Y_{\leq t}$ 
into future actions $A_{\geq t}$ 
and the \textit{environment channel}, 
$\mathbb{P}(O_{> t}|A_{\geq t},Y_{\leq t})$ that maps the joint histories 
$Y_{\leq t}$ and future actions $A_{\geq t}$ into future observations $O_{> t}$. 
An agent's preferences over the future dynamics $Y_{>t}$
are defined via its reward function 
$R:\mathbf{O} \times \mathbf{A} \mapsto \mathbb{R}$. 
The \textit{optimal policy} of the agent $\pi^{*}(o_{\leq t},a_{<t})$ 
maximizes the expected reward
$\mathbb{E}[R(Y_{>t})|Y_{\leq t});\pi^{*}))]$. 
We restrict our attention to environment channels 
and agent policies that generate a stochastic process 
$\mathbb{P}(Y_{>t}|Y_{\leq t})$ that is \textit{ergodic stationary},
i.e., processes for which the probability 
of every bi-sequence $(a_t,o_{t+1},..,a_{t+L},o_{t+1+L})$ 
of finite length $L\in \mathbb{Z}^{+}$, 
for some $L$ and all $t$ is time-invariant,
which can be reliably estimated from empirical data.

Formally, POMDPs suppose a hidden Markov process 
$\mathbb{P}(S_{>t}|S_t,A_t)$, 
with realizations $s_t\in\mathbf{S}$,
and observations emitted 
through the action-conditional probability $\mathbb{P}(O_{t+1}|S_t,A_t)$ \citep{kaelbling1998planning}. 
The causal relationship between the observed process 
and the hidden states implies 
that the mutual information $I[O_{t+1};S_t,A_t]$ 
between the \textit{generator state} $S_t$ 
and current action $A_t$ (jointly)
and the next observation $O_{t+1}$ 
is at least as great as that achieved by any competing
representation of the history.

This next-step sufficiency 
is extended to the infinite 
due to the recursive nature of generator.
$$I[O_{> t};S_t,A_t]\geq I[O_{> t};Y_{\leq t},A_t].$$

\subsubsection{Belief and Predictive State Representations}
A typical approach to planning in POMDPs
assumes the agent has access to 
$\mathbb{P}(S_{>t}|S_t,A_t)$ and $\mathbb{P}(O_{t+1}|S_t,A_t)$
and uses it to construct the belief states 
$b_t=\mathbb{P}(S_t|Y_{\leq t})$ 
from the finite realizations $Y_{\leq t}$~\citep{kaelbling1998planning}. 
Belief states are computed recursively using Bayes' formula 
from an initial belief $b_0=\mathbb{P}(S_0)$ 
and give rise to the belief process 
$\mathbb{P}(B_{>t}|B_{\leq t},Y_{>t})$. 
The belief process is a sufficient statistic 
of the \textit{generator state} 
when $$I[O_{> t};s_t,a_t]= I[O_{> t};b_t,a_t],$$ 
and is said to be \textit{asymptotically synchronized} 
when  $$\lim\limits_{L\to\infty}H[S_t|Y_{\leq t,>t - L}]=H[S_t|b_t]=0,$$ 
where $H$ is the conditional-entropy function. 
When $I[O_{t+1};S_t,A_t]>I[O_{t+1};Y_{\leq t},A_t]$,
the generator states contain more information about the future observable 
than the complete history of observations $Y_{\leq t}$,
implying absence of \textit{asymptotical synchronization} \citep{crutchfield2010synchronization}
and that belief states are only sufficient statistics 
of the history $Y_{\leq t}$ such that
$$I[O_{> t};S_t,A_t]>I[O_{> t};b_t,A_t]=I[O_{> t};Y_{\leq t},A_t].$$

The PSR approach relaxes the requirement that we know the underlying generative model, instead constructing representation using 
the outputs of the predictive model, 
$$\mathcal{M}_L=\{\mathbb{P}(O_{>t, \leq t + L}|Y_{\leq t},A_{\geq t, < t + L}=q_1),..,\mathbb{P}(O_{>t, \leq t + L}|Y_{\leq t},A_{\geq t, < t + L}=q_n)\},$$
of the next $L$ observations,
conditioned on the next $L$ actions $A_{\geq t, < t + L}$ (the test) sampled from the set of feasible $L$-length action sequences $\mathbf{Q_L}=\{q_1,..,q_n\}$~\citep{littman2002predictive}. 
By $\mathcal{M}$, we indicate
the collections of predictive models 
for all $L\in \mathbb{Z}^{+}$. 
Each model $\mathcal{M}_L$ is a sufficient statistic 
of the $L$-length future observations $O_{>t, \leq t + L}$, 
and the complete collection $\mathcal{M}$ 
is a sufficient statistic of the future observations $O_{> t}$, i.e.,  
$I[O_{> t};\mathcal{M},A_t]=I[O_{> t};Y_{\leq t},A_t]$. 
Traditionally, PSRs are constructed 
using a linear model that enables approximate solutions 
by assuming that the infinite-dimensional 
\textit{system dynamics matrix}\footnote{A matrix of probabilities of all combinations of histories and tests} has finite rank~\citep{singh2004predictive}. 

\subsubsection{Causal States Representations}
We now  show how the \textit{causal states representation} connects to PSRs by allowing the definition of a formal equivalence
between the latent generator states 
and the causal states reconstructed from history. 
As in the PSR framework, causal states 
depend on a predictive model of the observation process. 


\begin{definition}
\label{def:causal_states}
\citep{crutchfield1989inferring,shalizi2001computational}
The \textbf{causal states}
 of a stochastic process are partitions $\sigma\in\mathbb{S}$  
of the space of feasible pasts $Y_{\leq t}$ 
induced by the causal equivalence $\sim_{\epsilon}$:
\begin{equation}
\label{causal_eq}
 y_{\leq t}  \sim_{\epsilon} y_{\leq t}^{'} \iff \mathbb{P}(Y_{>t}|Y_{\leq t}=y_{\leq t})=\mathbb{P}(Y_{>t}|Y_{\leq t}=y_{\leq t}^{'}).
\end{equation}
Which implies:
\begin{equation}
 \mathbb{P}(Y_{>t}|S_t=\sigma_i)=\mathbb{P}(Y_{>t}|Y_{\leq t}=y_{\leq t})\quad \forall y_{\leq t}\in \sigma_i,
\end{equation}
\end{definition}
where $S_t$ is the variable denoting the causal state at time $t$, 
overwriting the definition in \cref{sec:decision_proc} 
of the unknown ground truth state.
Since all histories belonging to the same equivalence class 
predict the same (conditional) future, 
the corresponding causal state can be used 
to compress the information content of those histories.

In the computational mechanics literature, causal state models are usually called $\epsilon$-machines and are formally defined as:
\begin{definition} The \textbf{$\epsilon$-machine}
 of a stochastic process $\overleftrightarrow{Y}$ is given by the tuple $\epsilon = \langle \mathcal{S},\mathcal{Y},\mathcal{T}\rangle$
where $\mathcal{S}$ is the discrete alphabet of causal states, $\mathcal{Y}$ the discrete alphabet of observation and $\mathcal{T}$ is a set of observation conditional state-to-state transition matrices~\citep{crutchfield1989inferring,shalizi2001computational}.
\end{definition}
To summarize, the \textbf{$\epsilon$-machine} of a stochastic process 
is the minimal unifilar hidden Markov model 
able to generate its empirical realizations. 
The hidden states of an $\epsilon$-machine are called 
the causal states of the process, 
and correspond to partitions of the process history.
Furthermore, when the $\epsilon$-machine has finite causal states, 
it is possible to represent it as a labeled directed graph 
$\mathcal{G}=(\mathcal{S},T^{(o|a)}_{ij})$, 
with causal states as nodes and action-observation conditional transitions as edges. 
Because of the unifilar property, once an agent has perfect knowledge of the current causal states, 
the information in the future action-observation process are sufficient 
to uniquely determine the future causal states. 
This property can be exploited by multi-step planning algorithms 
which need not keeping track of potential stochastic transitions between the underlying states. This enables a variety of methods that are otherwise amenable only to MDPs like Dijkstra's algorithm. Empirical  support for this approach  can be found  in  \cref{app:planning}. 

It can be demonstrated \citep{shalizi2001computational} 
that the partition induced by $\sim_{\epsilon}$ 
is the coarsest possible and generates 
the minimal sufficient representation across the model class. 
In other words, this partition has the lowest cardinality 
while being capable of modeling all possible futures of the given system.
Sampling of new symbols in the sequence induces 
the creation of new histories and consequently new causal states. 
Because of this mapping from histories to states,
the resulting hidden Markov model is \textit{unifilar}, which we define below.

\begin{definition}
\label{def:hmm}
\citep{shalizi2001computational}
A \textbf{unifilar hidden Markov model} is a HMM 
whose state transition probability $\mathbb{P}(S_{t+1}|S_{t})$ 
is deterministic if conditioned on the output symbol, 
i.e., $ H\lbrack S_{t+1}|Y_{t+1},S_{t}\rbrack = 0$. 
\end{definition}
With explicit reference to the joint input-output history, 
the state transition dynamics are governed
by input-conditional transition matrices 
$\mathbf{T}^{o|a}\in \mathcal{T}$ with elements:
\begin{equation}
\label{cond_transit}
\mathcal{T}_{ij}^{o|a}=\mathbb{P}(\mathbf{S}_{t+1}=\sigma_j,O_{t+1}=o|\mathbf{S}_t=\sigma_i,A_t=a).
\end{equation}
All POMDPs can be captured by this model,
but the causal states are not necessarily finite 
in continuous state-action settings, and in environments with non unifilar generators.
We further explore the infinite causal state setting 
in the empirical results below in \cref{sec:doom} and \cref{sec:atari}.
Since the causal states are defined over histories of joint symbols, 
the causal state model is unifilar with respect to the joint variable $A_{t},O_{t+1}$, i.e.,
the transitions between states are deterministic
once the next action and observable have been sampled 
or $ H\lbrack S_{t+1}|A_{t},O_{t+1},S_{t}\rbrack = 0$. 
Unifilarity implies that the recurrent dynamics of the causal states 
are fully specified by the state-action-conditional 
symbol emission probability $\mathbb{P}(Y_{t+1}|S_{t},A_{t})$ 
and the action-symbol-conditional causal state emission probability 
$\mathbb{P}(S_{t+1}|Y_{t+1},A_{t},S_{t})$. 
As a consequence, knowledge of the current causal states $S_t$
and of the future action-observation sequence $O_{>t},A_{\geq t}$ 
induces a deterministic sequence of future causal states $S_{>t}$, 
 $H(S_{>t}|O_{>t},A_{\geq t},S_t)=0$.
 
\subsubsection{Stochastic Processes with Finite Causal States}
When the joint process admits a finite causal state representation, 
it is called a \textit{finitary} stochastic process 
which has multiple theoretical implications. 
In discrete stochastic processes with finite actions, 
finite-symbol alphabets, and finite memory of length $k$,
the causal states are always finite, 
with a worst case scenario in which each sub-sequence of length $k$ 
belongs to a distinct causal state forming a $k$-length Markov model \citep{shalizi2001computational}. 
When the causal states are finite, 
they are also unique up to isomorphisms \citep{shalizi2001computational} 
and always generate a stationary stochastic process. 
If the underlying generator is non-unifilar,
the causal states can be infinite and have the same information content 
as the potentially non-synchronizing belief states of the generator,
and the belief states defined over the causal states 
always asymptotically synchronize to the actual 
causal states \citep{crutchfield2010synchronization}. 

We first focus on partially-observable environments 
with discrete causal states, and either continuous or discrete observations. 
For continuous observations it is not possible to derive 
generic conditions that imply discrete or finite causal states. 
Therefore, the existence of discrete latent states 
has to be directly assumed or derived from alternative assumptions, 
like the existence of finite latent discrete variables 
underlying each continuous observation. 
When a memory-less map from continuous observation 
to latent discrete variables exists, 
the causal states of the revealed continuous variable process coincide 
with those defined over the underlying discrete variables up to isomorphisms.

\subsection{State Abstractions and Bisimulation}
State abstractions have been studied as a way to distinguish relevant from irrelevant information~\citep{li2006stateabs} in order to create a more compact representation for easier decision making and planning. \citet{bertsekas1989bounds,Roy06sabounds} provide bounds for approximation errors for various aggregation methods, and \citet{li2006stateabs} discuss the merits of \textit{abstraction discovery} as a way to solve related MDPs.

Bisimulation relations are a type of state abstraction that offers a mathematically precise definition of what it means for two environments to `share the same structure'~\citep{larsen1989bisim,Givan2003EquivalenceNA}. We say that two states are bisimilar if they share the same expected reward and equivalent distributions over the next bisimilar states. 
For example, if a robot is given the task of washing the dishes in a kitchen, changing the wallpaper in the kitchen doesn't change anything relevant to the task. One then could define a bisimulation relation that equates observations based on the locations and soil levels of dishes in the room and ignores the wallpaper. These relations can be used to simplify the state space for tasks like policy transfer \citep{castro2010using}, and are intimately tied to state abstraction. For example, the \textit{model-irrelevance abstraction} described by \citet{li2006stateabs} is precisely characterized as a bisimulation relation.
\begin{definition}[\citealt{Givan2003EquivalenceNA}]
\label{def:bisimulation}
Given an MDP $\langle S, A, P, R\rangle$, an equivalence relation 
$E:S\times S\mapsto \{0,1\}$ is a bisimulation relation 
if for all pairs of states $s_1,s_2\in S$ where the states are bisimilar, 
$s_1 E s_2$, the following properties hold:
\begin{enumerate}
    \item $R(s_1,a)=R(s_2,a)$
    \item $P(s'|s_1,a) = P(s'|s_2,a), \quad \forall s'\in S/E$
\end{enumerate}
\end{definition}
where $S/E$ denotes the partition of $S$ into $E$-equivalence classes.
\subsection{Causal Inference Using Invariant Prediction}
\label{sec:ci_with_icp}
We define the assumptions in the causal inference setting from \citealt{peters2016icp} and relate it to POMDPs in \cref{sec:bisim_ip_connections}. 
Causal inference considers the setting where there are different experimental conditions $e\in\mathcal{E}$ and different i.i.d. samples of $(X^e,Y^e)$ from each environment. $X^e\in\mathbb{R}^p$ is the observed data and $Y^e$ the target variable, and only a subset $S^*\subset \{1,...,p\}$ of the variables in $X^e$ have a causal effect on $Y^e$, leading to the following relationship:
\begin{equation*}
        \forall e\in\mathcal{E}, Y^e=g(X^e_{S^*},\epsilon^e),
\end{equation*}
where $\epsilon^e$ is environment specific noise independent of $X^e_{S^*}$, $X^e_{S^*}$ are the variables from  $X^e$ with indices in  $S^*$, and $g$ a real-valued function. From this formulation we can see that we want to find an invariant model  $g$ that persists across all experiments  $e\in\mathcal{E}$.

\citet{peters2016icp} first introduced an algorithm, Invariant Causal Prediction (ICP), to find the \textit{causal feature set} $S^*$, the minimal set of features which are causal predictors of a target variable, by exploiting the fact that causal models have an invariance property~\citep{pearl2009do,scholkopf2012causal}. \citet{arjovsky2019irm} extend this work by proposing invariant risk minimization (IRM, see \cref{eq:irm}), augmenting empirical risk minimization to learn a data representation free of spurious correlations. They assume there exists some partition of the training data $\mathcal{X}$ into \textit{experiments} $e \in \mathcal{E}$, and that the model's predictions take the form $Y^e = \mathbf{w}^\top \bm{\phi}{(X^e)}$. IRM aims to learn a representation $\bm{\phi}$ for which the optimal linear classifier, $\mathbf{w}$, is invariant across $e$, where optimality is defined as minimizing the empirical risk $R^e$. We can then expect this representation and classifier to have low risk in new experiments $e$, which have the same causal structure as the training set.
\begin{equation}\label{eq:irm}
\begin{split}
    &\min_{{\begin{subarray}{l}\bm{\phi}: \mathcal{X} \rightarrow \mathbb{R}^d\\
    \mathbf{w} \in \mathbb{R}^d \end{subarray}}} \sum_{e \in \mathcal{E}} R^e(\mathbf{w}^\top \bm{\phi}(X^e)) \\
    &\text{ s.t. } \mathbf{w} \in \underset{\bar{\mathbf{w}} \in \mathbb{R}^d}{\text{arg min }} R^e(\bar{\mathbf{w}}^\top \bm{\phi}(X^e)) \quad \forall e \in \mathcal{E}.
\end{split}
\end{equation}

The IRM objective in \cref{eq:irm} can be thought of as a constrained optimization problem, where the objective is to learn a set of features $\phi$ for which the optimal classifier in each environment is the same. Conditioned on the environments corresponding to different interventions on the data-generating process, this is hypothesized to yield features that only depend on variables that bear a causal relationship to the predicted value. Because the constrained optimization problem is not generally feasible to optimize, \citet{arjovsky2019irm} propose a penalized optimization problem with a schedule on the penalty term as a tractable alternative.

In order for invariant prediction to find a causal feature set, certain assumptions about the training data are needed. We require access to \textit{do-interventions}~\citep{pearl2009do} or \textit{noise-interventions}~\citep{eberhardt2007interventions}. Do-interventions are direct interventions that change the value of specific variables in $X$. Noise interventions are a form of ``soft" intervention that changes the distribution of a variable. A third form of intervention are \textit{simultaneous noise interventions}, which are useful in settings where it is not possible to only affect one variable at a time.
In an MDP, we can intervene on variables in the state space through actions~\citep{scholkopf2019causalityml}, but cannot guarantee that an action only affects one variable. 
In the  linear setting, using only simultaneous noise interventions can be sufficient for identifiability of the causal feature set $S^*$~\citep{peters2016icp}. We will use this result to explain under what assumptions causal states are causal feature sets.

\section{Connections to Bisimulation \& Invariant  Prediction}
\label{sec:bisim_ip_connections}
To simplify the notation when showing connections 
to bisimulation~\citep{ferns2004bisimulation}, 
we fold the reward into the observation seen by the agent,
i.e., $O:=(O,R)$ in the task-specific setting. 
We first connect this task-specific version of causal states 
to trajectory equivalence, an equivalence between states based on equivalent trajectories conditioned on action sequences (formally defined in \cref{def:traj_equivalence}),
and use Theorem 2.6 in \cite{castro2009equivalence}
to show that it is also a bisimulation. 
For notational convenience we introduce the following definition.
\begin{definition}[MDP of Histories of Observations (HOMDP)]
Given a POMDP \\
$\langle S, A, P, R, O, \gamma \rangle$ with state space $S$, 
action space $A$, reward function $R$, observation space $O$, 
transition distribution $P$, and discount factor $\gamma$, 
we can form an MDP $\langle \{O\}_0^t, P', A, R, \gamma \rangle$ 
with histories of observations $O_{\leq t}$ as states
and transition distribution $P'$. 
We denote this the corresponding \textbf{HOMDP} of our POMDP.
\end{definition}
We prove that causal states are a reduction (coarser bisimulation) of the HOMDP in \cref{thm:causal_states_bisimilar}\footnote{This is similar to the belief state MDP except the states 
are now learned causal states and the beliefs are Dirac functions.}. 
Any optimal policy on the causal states MDP 
generalizes to an optimal policy on the HOMDP.

We also show that we can relax our assumptions 
to the continuous state space setting,
where causal states are infinite by removing the discretization step. 
In this setting, the continuous representation 
can be connected to bisimulation metrics~\citep{ferns2011contbisim}, as shown in \cref{thm:bisim_bound}. 
Bisimulation metrics upper bound the optimal 
value function difference between two states (\cref{thm:value_lipschitz_bound}). 
By using the trick
of joining the HOMDP and causal state MDP 
into a single super-MDP 
(where we give up the irreducibility assumption common to the RL setting, i.e.,
we can no longer reach every state $s_i$ from any other state $s_j$),
we can bound the distance between the optimal value function 
of the causal state MDP to the HOMDP,
showing that the optimal value function has bounded Lipschitz constant 
with respect to bisimulation metrics, shown in \cref{thm:total_lipschitz}. 
With this result, we can lower bound the value function learned 
using causal states with respect to the optimal value function of the MDP 
defined by the history of observations (\cref{thm:valuefn_bound}). 

Finally, we draw a connection between the causal states objective and invariant prediction for causal inference. 
Given certain requirements on the diversity of data seen at training time, namely an intervention on every underlying variable in our MDP (as discussed in \cref{sec:ci_with_icp}, causal  states will learn the causal feature set. 

\subsection{Causal States are a Bisimulation}
\label{sec:causal_states_bisim}
We start with showing the equivalence between causal states and bisimulation.
\begin{definition}[\citealt{castro2009equivalence}]
\label{def:traj_equivalence}
Given an MDP and a partitioning of the state space 
into disjoint subsets via $\Psi:S\mapsto \Psi(S)$, 
states $s_1,s_2\in S$ are $\Psi$-\textbf{trajectory equivalent} 
if and only if $\Psi(s_1)=\Psi(s_2)$ and for 
any finite action sequence $a_{\geq t}\in A_{\geq t}$ 
and reward-state trajectory $\alpha\in (\mathbb{R}\times \Psi(S))^*$ 
conditioned on that action sequence,
\begin{enumerate}
\item For any $a\in A$, $R(s_1,a)=R(s_2,a)$.
\item $\mathbb{P}(\alpha|\Psi(s_1),a_{\geq t})=\mathbb{P}(\alpha|\Psi(s_2),a_{\geq t}).$
\end{enumerate}
\end{definition}
Here $\Psi^*$ is the fixed point of the iterates $\Psi^{(n)}$, 
which is equivalent to bisimulation. 
A more thorough discussion of this relation 
can be found in~\citet{castro2009equivalence}.

\begin{restatable}[Causal States  are Bisimilar]{thm}{causalstatesbisimilar}
\label{thm:causal_states_bisimilar}
If two observation sequences belong to the same causal state 
as per \cref{def:causal_states}, 
they are also bisimilar as per \cref{def:bisimulation}.
\end{restatable}
Proof in \cref{app:causal_states_bisimilar}. 
By showing the state aggregation defined by causal states is also a bisimulation, 
we prove that optimal value functions on the causal state MDP are also optimal for the HOMDP. 

\subsection{Lower Bounds on the Optimal  Value Function via Bisimulation Metrics}
Next, we bound the optimal value function difference 
between any two states under an approximate causal state representation \ 
and taking into account model error.  
We define a bisimulation metric:

\begin{definition}[\citealt{ferns2004bisimulation}]
\label{def:bisim_metric}
Given an MDP $\langle S, A, P, R, \gamma \rangle$, we define $$F(d)(s_1,s_2)=\max_{a\in A} ((1-\gamma)|r_{s_1}^a-r^a_{s_2}|+\gamma W_d (P_{s_1}^a,P_{s_2}^a)),$$ 
where $W_d$ is the Wasserstein distance between probability distributions 
and $d$ a pseudometric in a space of pseudometrics $\mathcal{D}$. 
$F$ has a least fixed point $\Tilde{d}$, which is our bisimulation metric.
\end{definition}

This bisimulation metric $\Tilde{d}$ bounds 
the optimal value function difference between two states $s, s'$. 

\begin{restatable}[Theorem 3.20 in \citealt{ferns2011contbisim}]{thm}{valuelipschitzbound}
\label{thm:value_lipschitz_bound}
Let $V^*$ be the optimal value function for MDP $M$ 
with discount factor $\gamma\in [0,1)$. 
Then $V^*$ is Lipschitz continuous with respect to $\Tilde{d}$ 
with Lipschitz constant $\frac{1}{1-c}$ where $\gamma\leq c$,
$$|V^*(s) - V^*(s')|\leq \frac{1}{1-\gamma}\Tilde{d}(s,s').$$
\end{restatable}
This result is achieved using the Banach fixed point theorem 
and proves Lipschitz continuity of the optimal value function 
with respect to bisimulation metrics.

Our representation can be upper bounded by bisimulation metrics 
with an approach similar to \citet{gelada2019deepmdp}. 
We define Lipschitz constants for the transition and reward functions 
of an MDP with metric $d_s$ for the state space, 
Wasserstein distance for probability distributions over transitions, 
and $L_1$ distance for reward:
\begin{equation*}
\begin{split}
    L_D&:=\sup_{\substack{s,s'\in S \\ a\in A}}\frac{W_d(P(s,a), P(s',a))}{d_s(s,s')}, \\ 
    L_R&:=\sup_{s,s'\in S} \frac{|R(s) - R(s')|}{d_s(s,s')}.
\end{split}
\end{equation*}

\begin{restatable}[]{thm}{bisimbound}
\label{thm:bisim_bound}
Let $M$ be the HOMDP representing a POMDP whose states 
are the history of observations $O_{\leq t}$, 
and $\bar{M}$ be the $L_D$-$L_R$-Lipschitz MDP 
with $\mathcal{L}_d$ and $\mathcal{L}_r$ representing the losses. 
Let $\phi$ denote the encoder that
takes in the history of observations 
and outputs the approximate causal state. 
The bisimulation distance 
$\Tilde{d}:O_{\leq t}\times O_{\leq t}\mapsto \mathbb{R}$ in $M$ 
can be upper bounded by $L_2$ distance in $\bar{M}$ as
\begin{equation*}
\begin{split}
    \Tilde{d}(o^1_{\leq t},o^2_{\leq t})\leq \frac{(1-\gamma)L_R}{1-\gamma L_D}W_d\big(\phi(o^1_{\leq t}),\phi(o^2_{\leq t})\big) \\
    + 2\bigg(\mathcal{L}^\infty_r +\gamma \mathcal{L}^\infty_d \frac{L_R}{1-\gamma L_D}\bigg),
\end{split}
\end{equation*}
\end{restatable}
where $\mathcal{L}_{\text{x}}^\infty:=\sup_{s\in S,a\in A} \mathcal{L}_{\text{x}}$.
Proof found in \cref{app:bisim_bound}.

\cref{thm:value_lipschitz_bound} and \cref{thm:bisim_bound} 
can be combined to give an overall bound (\cref{thm:total_lipschitz}) 
on the smoothness of the optimal value function 
that depends on characteristics of the learned causal state representation.

\begin{restatable}[]{thm}{totallipschitz}
\label{thm:total_lipschitz}
Given a HOMDP $M$ and corresponding learned causal state MDP $\bar{M}$, 
the Lipschitz constant of the optimal value function of a HOMDP $M$ 
is upper bounded by $L_2$ distance in $\bar{M}$ as follows,
\begin{equation*}
\begin{split}
|V^*(o^1_{\leq t}) - V^*(o^2_{\leq t})|\leq \frac{L_R}{1-\gamma L_D}W_d\big(\phi(o^1_{\leq t}),\phi(o^2_{\leq t})\big) \\
    + 2\bigg(\mathcal{L}^\infty_r +\gamma \mathcal{L}^\infty_d \frac{L_R}{1-\gamma L_D}\bigg).
\end{split}
\end{equation*}
\end{restatable}

Using this fact that the optimal value func tion is Lipschitz, 
we can write a lower bound for the optimal 
state-action value function in our causal state MDP. 
Given the above bound on the optimal value function, 
we denote its Lipschitz constant $L_{V^*}.$
\begin{restatable}[]{thm}{valuefnbound}
\label{thm:valuefn_bound}
Given a HOMDP $\mathcal{M}$ with states $O_{\leq t}$,
where $0\leq t \leq H$ and causal state MDP $\hat{\mathcal{M}}$ 
with states $\phi:O_{\leq t}\mapsto \hat{S}$. 
The suboptimality of the state-action value function 
corresponding to the optimal policy in the causal state MDP $\hat{\pi}$ is bounded as follows:
\begin{equation}
Q^*(o_{\leq t}, a) - Q^{\hat{\pi}^*}(\phi(o_{\leq t}), a)\leq 2\frac{\mathcal{L}^\infty_r + \gamma L_{V^*}\mathcal{L}^\infty_d}{1-\gamma}.
\end{equation}
\end{restatable}
Proof in \cref{app:valuefn_bound}. 
This connection to bisimulation shows that 
there exists an optimal solution to the causal state objective and 
that we approach it by minimizing losses $\mathcal{L}_d^\infty$ and $\mathcal{L}_r^\infty$, giving a lower bound on the optimal state-action value function of the causal state MDP. This maps how well we can perform on a downstream control task directly to the quality of our learned causal states representation.
This result now lends itself to a 
gradient-based learning method, 
detailed in \cref{sec:method}. 

\subsection{Causal States are a Causal Feature Set}
We now show how learning causal states connects to invariant prediction, and learns the causal  feature set under certain requirements. 
Invariant causal prediction aims to identify a set $S^*$ of causal variables such that a linear predictor with support on $S^*$ will attain consistent performance in next reward and  observation prediction under all  policies. In other words, ICP removes irrelevant variables from the input, just as state abstractions remove irrelevant information from the environment's observations.

We consider whether such a state abstraction can be obtained by ICP. Intuitively, one would then expect that the \textit{causal variables} should have nice properties as a state abstraction. The following result confirms this to be the case; a state abstraction that selects the set of causal variables from the observation space of a HOMDP will be a bisimulation.

\begin{assumption}[Environment Interventions]
\label{ass:interventions}
For a HOMDP $\langle \{O\}_0^t, P', A, R, \gamma \rangle$, let $S = S_1 \times \dots \times S_n$ be the underlying  variables  of the observations $O$. Each action $a \in A$ corresponds to a simultaneous noise-intervention on some subset of variables $S_i$. 
\end{assumption}

\begin{restatable}[Bisimulation is a Causal Feature Set]{thm}{causalstatesets}
\label{thm:causalstatesets}
Consider a HOMDP \\
$\mathcal{M} = \langle \{O\}_0^t, P', A, R, \gamma \rangle$. Let $\mathcal{M}$ satisfy \cref{ass:interventions}. Let $S_{RO} \subseteq \{1, \dots, k\}$ be the set of variables such that the reward $R(s,a)$ and future observations $O$ are a function only of $[s]_{S_{RO}}$ ($s$ restricted to the indices in $S_{RO}$). Then let $S = \textbf{AN}(RO)$ denote the ancestors of $S_{RO}$ in the (fully observable) causal graph corresponding to the transition dynamics of $\mathcal{M}$. Then the state encoder $\phi_S$ that produces abstraction $\phi_S(\{O\}_0^t) = [s]_S$ is a bisimulation, and therefore also produces \textit{causal  states}. 
\end{restatable}

\begin{figure}
    \centering
    \includegraphics[width=0.45\textwidth]{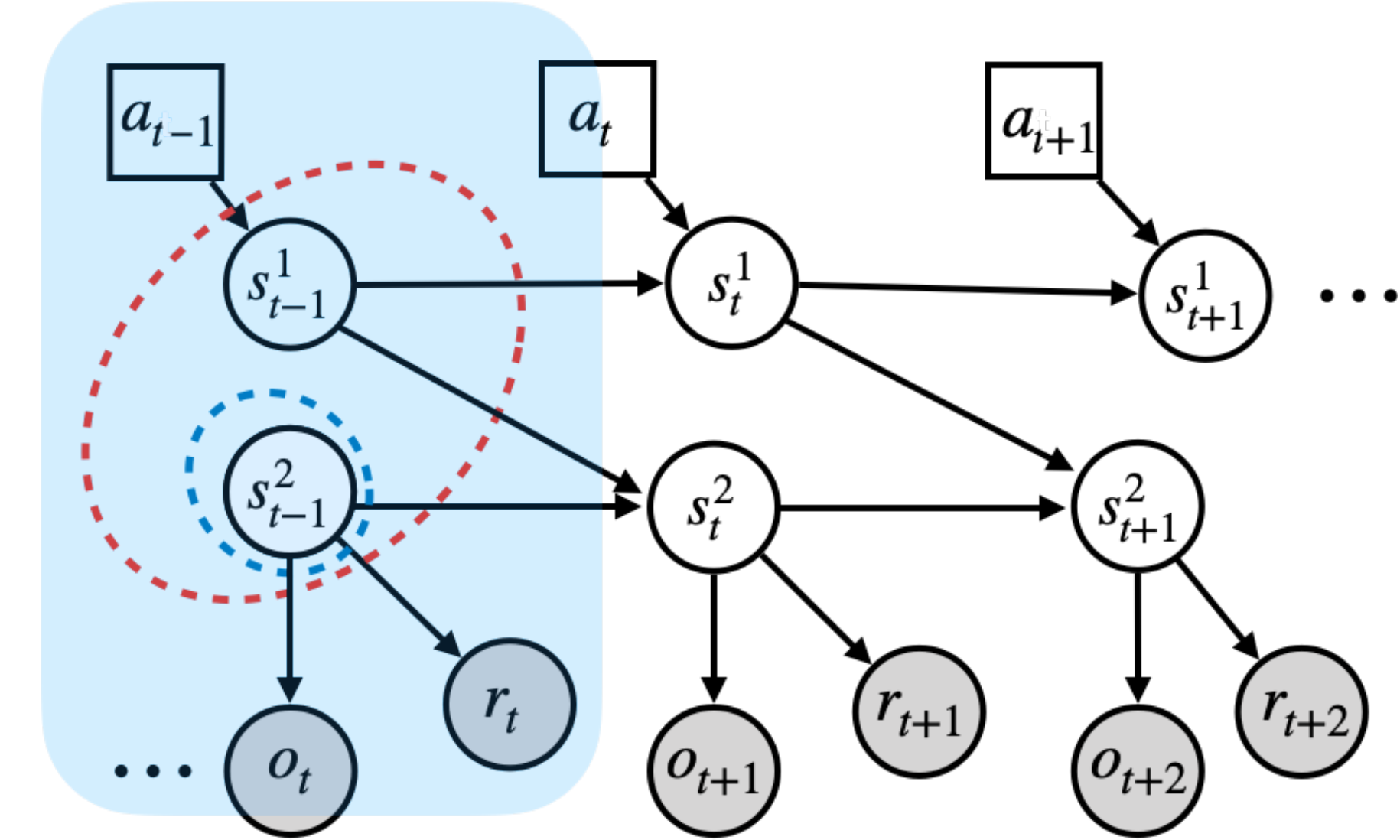}
    \caption{Graphical causal models with temporal dependence -- note that while $s^2$ (circled in blue) is the only causal parent of the reward and observation, because its next-timestep distribution depends on $s^1$, a bisimilar abstraction must include both variables. Shaded in blue: the graphical causal model of a POMDP with states $s=(s^1, s^2)$ when ignoring timesteps. }
    \label{fig:statgm}
\end{figure}

An important detail in the previous result, established in \cite{zhang2020invariant} is the bisimulation relation incorporates not just the parents of the reward and next observation, but also their ancestors. This is because in RL, we seek to model \textit{return} rather than solely rewards, which requires a state abstraction that can capture multi-timestep interactions. We provide an illustration of this in \cref{fig:statgm}. As a concrete example, we note that in the popular benchmark CartPole, only position $x$ and angle $\theta$ are necessary to predict the reward. However, predicting the return requires $\dot{\theta}$ and $\dot{x}$, their respective velocities. 

If interventions by actions seen at training time cover all variables that can be intervened on, then the causal feature set that is robust to any interventions on those variables at test time is identifiable. 
We can now use the result from \cref{sec:causal_states_bisim} to connect causal states to learning causal feature sets, as we've shown that causal states and causal feature sets both correspond to a bisimilar abstraction of the original HOMDP.

\section{Methods for Reconstructing Causal States from Empirical Data}
\label{sec:method}
In the previous sections, we introduced a class of stochastic processes 
with discrete or continuous outputs
that are optimally compressed by a finite-state hidden-Markov representation, 
called the causal state model of the process.
Existing methods either directly partition past sequences of length $L$ 
into a finite number of causal states via conditional hypothesis tests \citep{shalizi2004blind} 
or use Bayesian inference over a set of existing candidate states~\citep{strelioff2014bayesian}. 
Either method can be adapted to model a joint-process 
and consequently obtain the next-step conditional output 
by marginalizing out the action $A_t$, 
but do not extend to the real-valued measurement case described 
without strong assumptions on the shape of the conditional density function. 

Methods for learning a causal feature set generally require access to the underlying state variables, and are super-exponential in the number of variables~\citep{peters2016icp}, requiring hypothesis testing of every possible subset of variables to find the causal set. \cite{zhang2020invariant} exploits invariance across tasks to learn an invariant representation that corresponds to a bisimulation, but relies on the multi-task (and fully observable) setting. 
\textbf{We now propose a new approach to approximately reconstruct these causal states from empirical data.}

\subsection{Learning Sufficient Statistics of History with Recurrent Networks}
\label{rnn_history}
Recurrent neural networks (RNNs) are unifilar hidden Markov models 
with continuous states, where the transitions and state output probabilities 
are parameterized by differentiable functions~\citep{hinton2013rnn}. 
We use them to obtain recursively-computable 
high-dimensional sufficient statistics 
of the action-measurement joint process.
This representation is learned via a recurrent encoder 
$\phi: Y_{\leq t}\mapsto \hat{\mathbf{S}}$, 
a next step prediction network 
$f:  \hat{\mathbf{S}}\times\mathbf{A} \mapsto \mathbf{O}$,
and next step reward prediction network
$r:  \hat{\mathbf{S}}\times\mathbf{A} \mapsto \mathbb{R}$. 

We note that when $f(\hat{s}_t,a_{t})$ is maximally predictive 
of the subsequent observations (the future $O_{\geq t}$), 
$\hat{s}_t$ constitutes a sufficient statistic 
of the latent states $S_{>t}$. 
In practice, we estimate the continuous representations 
using the empirical realizations\footnote{With a small abuse of notation, we use the same convention of $Y_{>t}$, where the joint process $\{O,A\}$ has elements ($O_{>t},A_{\geq t}$).}  $(o_{>t},a_{\geq t})$
to learn a neural network that approximates end-to-end the maps $\phi$ and $f$ 
by minimizing the temporal loss through the following objectives:
\begin{equation}
\begin{aligned}
\label{rnn_enc_loss}
    \mathcal{L}_D&=\sum_t^{T}W\big(\mathbb{P}(O_{t+1}|o_{>t},a_{\geq t},a_{t}),f(\phi(o_{>t},a_{\geq t}),a_{t}) \big), \\
    \mathcal{L}_R&=\sum_t^{T}\big|R(O_{t+1}|o_{>t},a_{\geq t},a_{t}) - r(\phi(o_{>t},a_{\geq t}),a_{t}) \big|.
\end{aligned}
\end{equation}

After solving \cref{rnn_enc_loss} we use the neural networks parameterized by the optimal parameters $\phi^*, f^*$ to derive sufficient continuous representations to create discrete states that are refinements of the causal states. 

\begin{figure}[t]
    \centering
    \includegraphics[width=.7\linewidth]{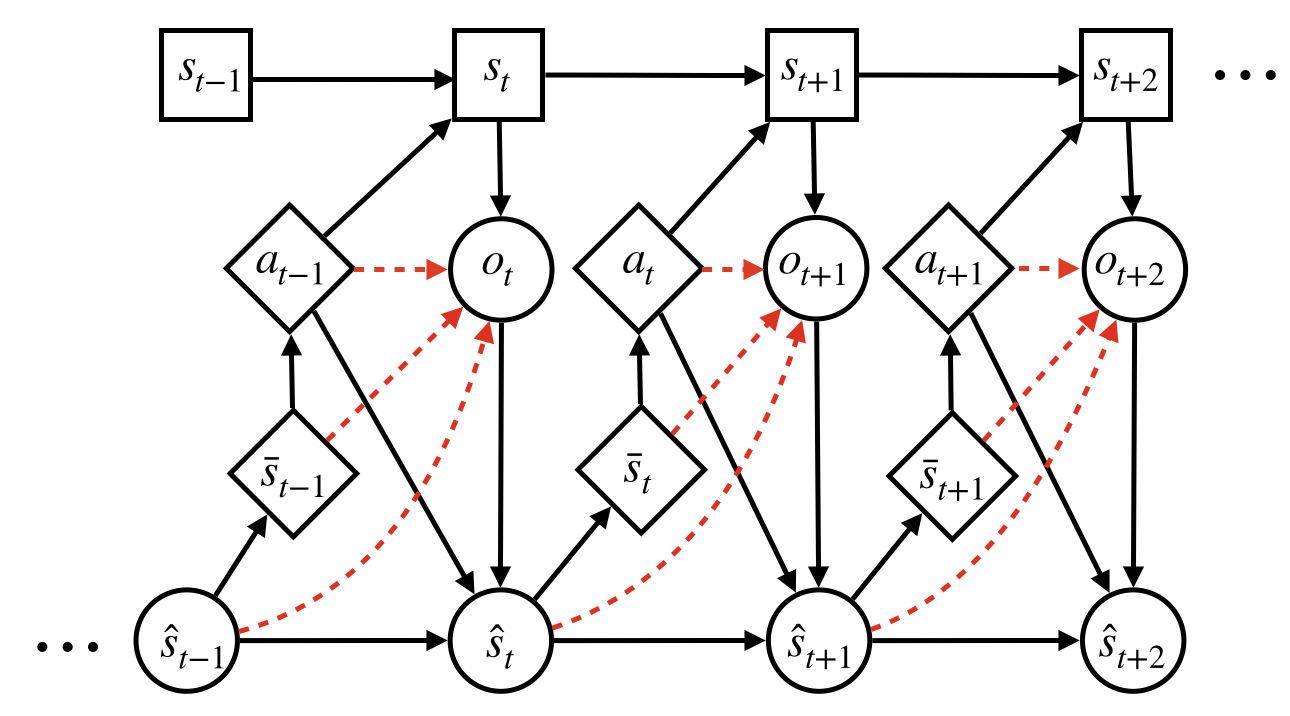} 
    \caption{ Graphical model of the joint action-measurement stochastic process.
    Black-arrows indicate causal relationships between random variables and 
    red-arrows indicate the predictive relationship between the action $a_t$,
    internal state $\bar{s}$ ($\hat{s}$) and the next measurement $o_{t+1}$. 
    Circular boxes indicate continuous variables.}
    \label{fig:envl}
\end{figure}

\subsection{Discretization of the RNN Hidden States}
Together with the unifilar and Markovian nature of transitions in RNNs,
the sufficiency of $\hat{s}_t $ implies
that there exists a function $\mathcal{D}^{s}: \mathbb{R}^{k} \mapsto \mathbf{S}$ 
that allows us to describe the causal states $s$ 
as a partition of the learned latent state $\hat{s}$ \citep{shalizi2001computational,crutchfield2010synchronization}.

We set up a second optimization problem using the trained neural network 
and the empirical realizations of the process $o$ 
to estimate the discretizer $\bar{\phi}:\mathbf{\hat{S}}\mapsto \mathbf{\bar{S}}$ with $|\mathbf{\bar{S}}| = |\mathbf{S}|$ 
and the new dynamics and reward prediction networks $\bar{f}:\mathbf{\bar{S}}\times A \mapsto \mathbf{O},\bar{r}:\mathbf{\bar{S}}\times A \mapsto \mathbb{R}$
that map the estimated discrete states into the next observable. 
We match the predictive behavior between the old network $\phi$
and the new networks $\Lambda(o_{\leq t},a_{t})=( \phi \circ \bar{\phi} \circ \bar{f})(o_{\leq t},a_{t}),( \phi \circ \bar{\phi} \circ \bar{r})(o_{\leq t},a_{t})$ 
that use discretized states $\bar{s}$ and corresponding prediction functions $\bar{f}, \bar{r}$ 
by minimizing the knowledge distillation \citep{hinton2015distilling} loss:
\begin{equation}
\label{drnn_loss}
    \min_{\substack{\bar{f},\bar{r},\bar{\phi}} }
    \sum_t^{T}\mathcal{L}_{\text{KD}}\big(f(\phi(o_{\leq t},a_{<t}),a_t),
    r(\phi(o_{\leq t},a_{<t}),a_t),\Lambda(o_{\leq t},a_{t}) \big).
\end{equation}

When \cref{drnn_loss} goes to 0 we have a sufficient discrete representation. Any discretization method can be used here, either learning-based such as VQ-VAE~\citep{vandenoord2017vqvae} and ternary tangent neurons~\citep{koul2018learning} or clustering-based, such as k-means. 
\cref{fig:envl} shows the stochastic process representing the environment and our learned states $\hat{S}$ and $\bar{S}$ and their interactions. In \cref{algo:causal_states} we present our method both with and without discretization, where $\mathcal{L}_Q$ is the Bellman update. We initialize the continuous causal state encoder $\phi$, optional discrete causal state encoder $\bar{\phi}$, $Q$ function, dynamics model $f$, reward model $r$, and replay buffer $\mathcal{D}$. At each timestep, we pass the causal state, generated by  $\phi$ or $\bar{\phi}$, to the $Q$ function to obtain the action that maximizes the estimated value, with $\epsilon$-greedy exploration. The agent steps in the environment, and saves that transition to the replay buffer. Then we sample a batch of data, and use it to update the encoder through the transition and reward losses $\mathcal{L}_D$ and $\mathcal{L}_R$, respectively. The optional discrete $\phi$ is similarly updated, and finally the $Q$ function is updated with the Bellman update.

\subsection{Equivalence of 1-Step Supervision to Infinite Future}
Causal states are, by definition, checking equivalence over all possible future trajectories. In practice, with reinforcement learning, we do not have access to these counterfactual trajectories, but only a single trajectory taken for a given history, usually. In this setting, supervision of the next step observation is sufficient to learn a causal state representation. To see why, note that full episode trajectories are stored in the replay buffer but sampled as independent transitions. If we  perform 1-step supervision (with observation-action histories) over all batches in the replay buffer, then each step in the episode is checked independently for equivalence. 

\begin{algorithm}[t]
\SetCustomAlgoRuledWidth{0.6\textwidth}
\SetAlgoLined
\KwResult{$\phi$, a causal state encoder}
$Q_\theta,\phi,f, r \gets Q_0, \phi_0, f_0, r_0$ \; \\
 $h_0 \gets 0$ \\
 $\mathcal{D} \gets \emptyset$ \; \\
 \For{episode=1,$M$}
 {
    \For{$t=1,T$}
    {
        $a_t,h_t \gets \argmax_{a\in A} Q(\phi(o_t,h_{t-1}))$\; \# Replace with $\bar{\phi}$ if using discretization \\
        $o_{t+1}, r_t \gets $ \texttt{step}$(o_t,a)$ \; \\
        \texttt{store}$(o_t,a_t,r_t,o_{t+1})$ in $\mathcal{D}$ \; \\
        Sample batch $B$ from $\mathcal{D}$ \; \\
        $f, r, \phi \gets \nabla_{f,\phi} [\mathcal{L}_D(B) ] + \alpha_R\nabla_{r,\phi} [\mathcal{L}_R(B) ]$\; \\
        $\bar{f}, \bar{r}, \bar{\phi} \gets \nabla_{\bar{f},\bar{r},\bar{\phi}} [\mathcal{L}_\text{KD}(B) ] $ \; \# Optional discretization step \; \\
        $Q_\theta \gets \nabla_{\theta} \mathcal{L}_{Q}(B)$\;
    }
 }
 \caption{Learning Causal States with DQN}
\label{algo:causal_states}
\end{algorithm}

\section{Experiments}
We consider three settings to learn approximate causal states through self-supervised learning and use these representations as input for reinforcement learning tasks defined over the domains. The first setting consists of grid worlds with discrete state-action spaces, which are simple enough that we know exactly the  natural causal states. We use this set of experiments to validate that our method successfully uncovers the correct set of causal states. We further show the benefit of discrete causal states  -- we can construct a graph of experienced transitions and run Dijkstra's algorithm to obtain a policy. 

We then expand to the rich observation setting: 1) a similar maze environment in VizDoom~\citep{Kempka2016ViZDoom}, and 2) Atari in OpenAI Gym~\citep{brockman2016openai} with flickering to make it partially observable, to show our method works in settings with infinite causal states. In the different environments we compare with different baselines -- in gridworlds we show how the causal state representations, both discrete and continuous, fare against value iteration representations using RNNs and explicit memory. We validate what our theoretical results show -- that causal states are a principled method for uncovering true latent states and therefore achieve optimal performance, whereas value iteration RNN methods do not. In the rich observation settings, we compare with state-of-the-art deep RL methods for learning belief states -- Deep Variational Reinforcement Learning (DVRL, \cite{igl2018deep}) and RNN-based method Deep Recursive Q-Networks (DRQN, \cite{hausknecht2015deep}).

\subsection{Implementation Details}
For the Gridworld experiments, the base forward model architecture is composed of a three-layer perceptron (MLP) encoding the observation $o_t$ and a single layer linear embedding for the action $a_{t-1}$. The outputs of the respective layers are concatenated and fed to a Gated Recurrent Unit (GRU) \citep{chung2014empirical}, and the output of the recurrent network $\hat{s}_t$ is concatenated with the embedding of $a_{t}$ and fed to a second MLP that outputs predictions for $o_{t+1}$. All the embeddings have 64 neurons except in layout 4 where we use 256 dimensional embeddings. 

For the discretization phase we use two approaches that lead to equivalent performance. In the first, gradient based method we use a discretization network composed of a Quantized-Bottleneck-Network~\citep{koul2018learning} with ternary tangent neurons that auto-encodes the continuous representation $\hat{s}_t$ generating the discrete variables $\bar{s}_t$, and a MLP decoder that uses the estimated discrete states for predicting the next observation $o_{t+1}$.
Both networks are trained with the RMSprop algorithm using cross-entropy loss for discrete observations and reconstruction loss for the continuous setting. The world-model is trained through supervised learning of the temporal loss, while the discretization network is trained via knowledge distillation using the soft outputs of the GRU decoder as targets. We experimented with both online and offline discretization and found the offline procedure to lead to more stable training.
Alternatively, we simply run k-means clustering on the continuous representation $\hat{s}_t$ and then use it as input to predict the next observation $o_{t+1}$. We choose the minimum $k$ leading to satisfactory performances for the next step prediction network.
For all the tasks we ran downstream evaluation of our learned representation with value iteration and compare with baselines. To approximate the value function we use a $2$-layer fully-connected DQN architecture separated by ReLU with a 64-dimensional hidden layer. The DRQN~\citep{hausknecht2015deep} baselines use the same architectures but they are trained end-to-end via reward maximization.

In practice, we found using ternary neurons~\citep{koul2018learning} for discretization to work best among the competing gradient based methods like straight-through estimator or VQ-VAE. 
To summarize, we first minimize $\mathcal{L}_{r}$ 
to obtain a neural model able to generate continuous sufficient statistics 
of the future observables of the process 
and subsequently minimize $\mathcal{L}_{d}$ 
to obtain a sufficient representation of the dynamical system 
that is a refinement of the original causal states.

For the VizDoom and ALE experiments we build upon Rainbow DQN~\citep{rainbow} and drop the discretization step to account for continuous causal states.

\subsection{Gridworlds}
\label{sec:grid}
\begin{figure}[h]
    \centering
    \includegraphics[width=0.3\textwidth]{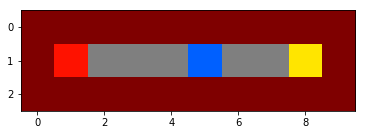}\hspace{1cm}
    \includegraphics[width=.3\textwidth,angle=90]{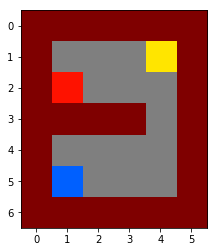}
    \caption{Visual representation of the layouts 1 and 2 used in the gridworld experiments. In \textit{Blue} the starting location, in \textit{Red} the key location, in \textit{Yellow} the final goal.}
    \label{fig:layout_map12}
\end{figure}

We create partially observable gridworld environments 
where the task is for the agent to first obtain the key, 
then pass through the door to obtain the final reward. 
The final state is unseen (i.e., the agent cannot pass through the door to reach it) \textit{if} the agent does not have the key.
The agent only knows it has the key if it remembers entering the state with the key, 
so without infinite memory this task is partially observable. 
At each time step the agent receives -0.1 reward, 0.5 reward for picking up the key, 
and a final reward of 1 for passing through the door.
We conduct experiments with two layouts shown in \cref{fig:layout_map12}.
The minimal memory requirement for solving the environment 
is given by the shortest path from the key to the final destination.

\begin{figure}[h]
    \centering
    Layout 1 \hspace{145pt}     Layout 2 \\
    \includegraphics[width=0.47\textwidth,trim=38 0 4 38,clip]{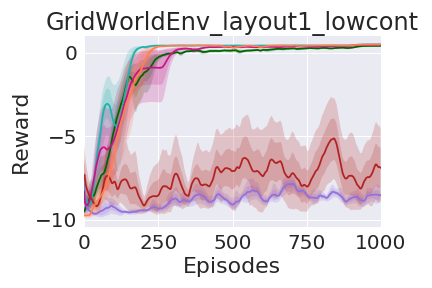}\hspace{-10pt}
    \includegraphics[width=0.47\textwidth,trim=38 0 4 38,clip]{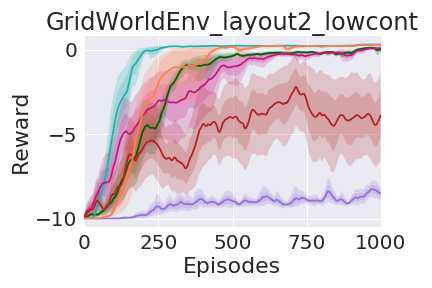}\\ 
    \includegraphics[width=1\textwidth,trim=0 0 0 1,clip]{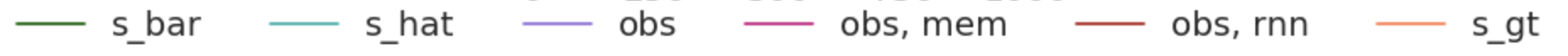}
    \caption{DQN training curves for Gridworlds across 10 seeds for Layouts 1 and 2 - Causal States $\hat{S}$ estimated using K-mean clustering. Y-axis is reward. Two levels of shading represent 1 and 2 standard deviations from the mean.}
    \label{fig:dqn_plots_l1}
\end{figure}

DQN on $\hat{S}$ is able to achieve optimal policy across all 10 random seeds with very low or zero standard deviation as seen in \cref{fig:dqn_plots_l1}, showing the stability of our learned $\hat{S}$. We expect $\hat{S}$ to perform as well as or better than $\bar{S}$, as $\bar{S}$ is distilled from $\hat{S}$ and therefore contains the same information or less. 
Using only current observation learns using a recurrent DQN (DRQN)~\citep{hausknecht2015deep}. We also compare against several baselines, including observation without memory (\texttt{obs}), both with explicit memory (\texttt{obs, mem}) and implicit memory in the form of an RNN (\texttt{obs, rnn}), and the ground truth state (\texttt{s\_gt}, which includes key information). 

\begin{table}[h]
\caption{ Results for Gridworlds. Reward obtained with tabular Q-learning, DQN, and DRQN with $\gamma=0.99$. 
 Models trained on 1000 episodes and evaluated on 100. 
 Numbers are mean, standard error across 10 random seeds. 
 First section is our method.
 Second is baselines on current observation, history of observations, and $S$. 
 Final section is using ground truth states. Note that tabular methods are deterministic and so std. is always 0.}
 \vspace{5px} 
    \centering
    \begin{tabular}{l|lll|lll}
    \toprule
    Method & Layout 1 & Layout 2 \\
    \midrule
     Tabular, $\bar{S}$ & $0.43 \pm 0.$ &  $0.01\pm 0.$ \\
     DQN, $\bar{S}$ & $\mathbf{0.50 \pm 0.005}$  & $-0.17 \pm 0.24$ \\
     DQN, $\hat{S}$ & $\mathbf{0.5 \pm 0.}$  & $\mathbf{0.30 \pm 0.}$ \\
     Dijkstra, $\bar{S}$ & \textbf{0.5, 0.} & \textbf{0.3, 0.}  \\
     \midrule
     DQN, $Y$  & $-9.46\pm 0.06$ &  $-9.48\pm 0.04$   \\
     DQN, $Y_{\leq t}$ & $-0.91\pm 0.95$ & $0.23\pm 0.05$ \\
     DRQN, $Y$ & $-9.75 \pm 0.07$ & $-5.63 \pm 1.18$ \\
     Tabular, $Y$ & $-9.40 \pm 0.$ & $-9.11 \pm 0.$ \\
     \midrule
     Tabular, $S_{gt}$ & $0.45 \pm 0.$ & $0.23\pm 0.$ \\
     DQN, $S_{gt}$ & $0.44\pm 0.01$ & $\mathbf{0.30\pm 0.003}$ \\
     Dijkstra, $S_{gt}$ & \textbf{0.5, 0.} &  \textbf{0.3, 0.} \\
     \bottomrule
     \end{tabular}
     \label{tab:gridworld_active}
     \vskip -0.2cm
\end{table}

A more extensive comparison can be found in \cref{tab:gridworld_active} with the inclusion of tabular methods. Surprisingly, DQN outperforms the tabular methods, perhaps because of insufficient exploration in the tabular regime. These results are using K-means for discretization, equivalent results using ternary neurons and gradient descent can be found in \cref{sec:add_gridworld}.

\subsection{Planning with Causal States}
\label{app:planning}
With a minimal sufficient unifilar model we can perform efficient planning
by representing it as a labeled directed graph 
$\mathcal{G}=(\mathcal{S},T^{(o|a)}_{ij})$, 
with causal states as nodes and action-observation conditional transitions as edges. 
Because of the unifilar property, once an agent has perfect knowledge of the current causal states, 
the information in the future action-observation process are sufficient 
to uniquely determine the future causal states. 
This property can be exploited by multi-step planning algorithms 
which need not keeping track of potential stochastic transitions between the underlying states, enabling a variety of methods that are otherwise amenable only to MDPs like Dijkstra's algorithm. We plan over our learned discrete representation by building a graph $G:=(V, E)$ where $V:=\{\bar{\mathbf{S}}\}$, $E:=\{(\bar{s}_i, \bar{s}_j); s_i,s_j\in\bar{\mathbf{S}}\}$, and $p(s_j|s_i,a)>0$ for $a\in\mathbf{A}$. 
We obtain the optimal policy for Layout 1 and 2 obtaining respectively 0.5 and 0.3 reward (\cref{tab:gridworld_active}).  
We derive $G$ and save high reward states seen during the rollouts as goal states. Then one can map the initial and final observations to a node in the graph and run Dijkstra's algorithm to find the shortest path as proposed in \cite{zhang2018composable}. Unlike value iteration, this requires no learning and no re-sampling of the environment by making use of the graph edges, where Q-learning only uses the nodes.

\subsection{Doom}
\label{sec:doom}
\begin{figure}[h]
    \centering
    \hspace{-0.15cm}
    \includegraphics[width=0.18\linewidth]{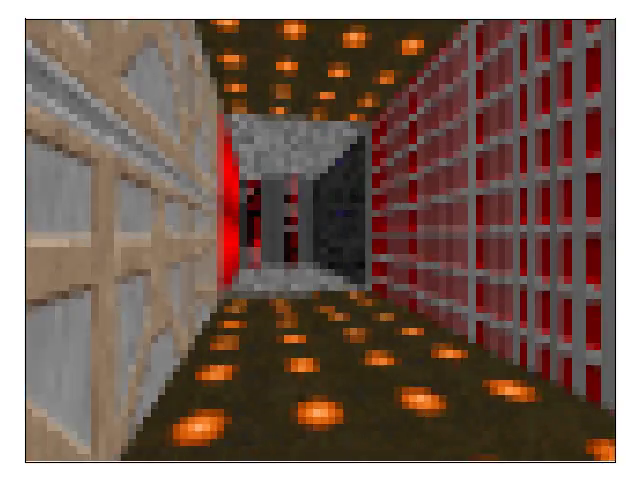}\hspace{-0.2cm}
    \includegraphics[width=0.18\linewidth]{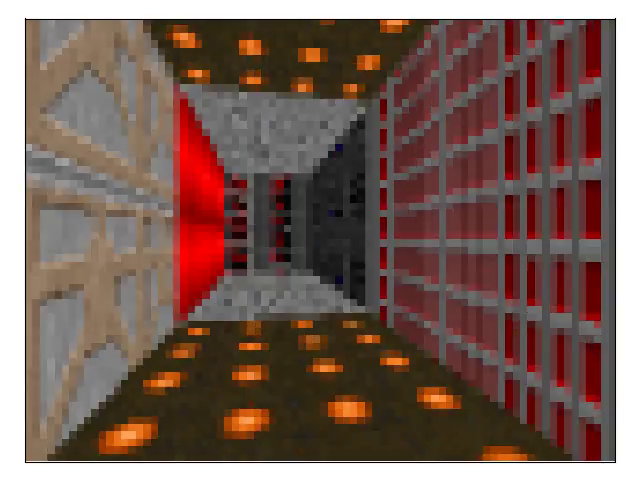}\hspace{-0.2cm}
    \includegraphics[width=0.18\linewidth]{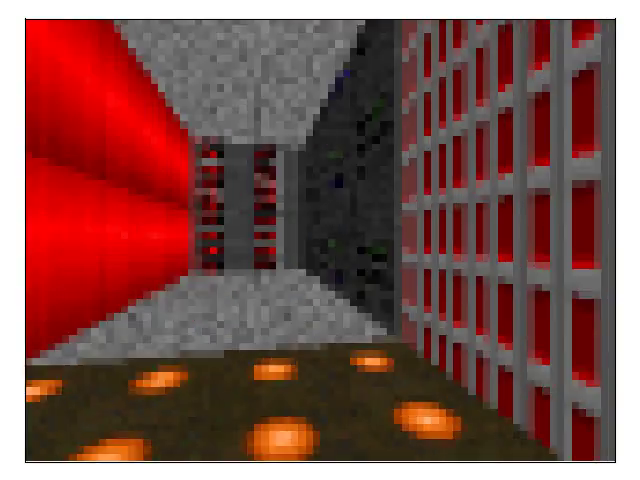}\hspace{-0.2cm}
    \includegraphics[width=0.18\linewidth]{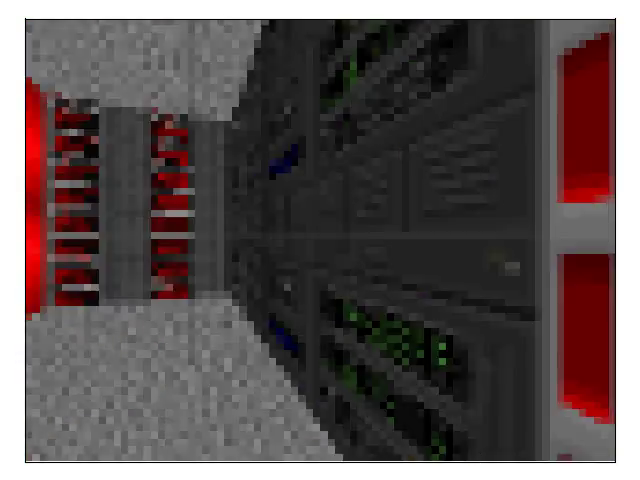} \\
    \includegraphics[width=0.18\linewidth]{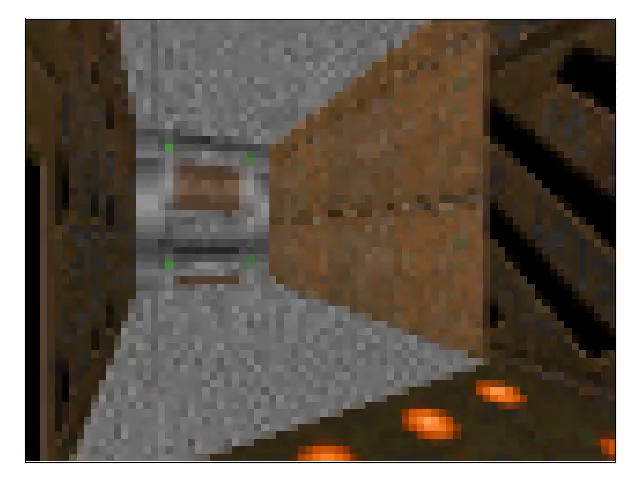}\hspace{-0.2cm}
    \includegraphics[width=0.18\linewidth]{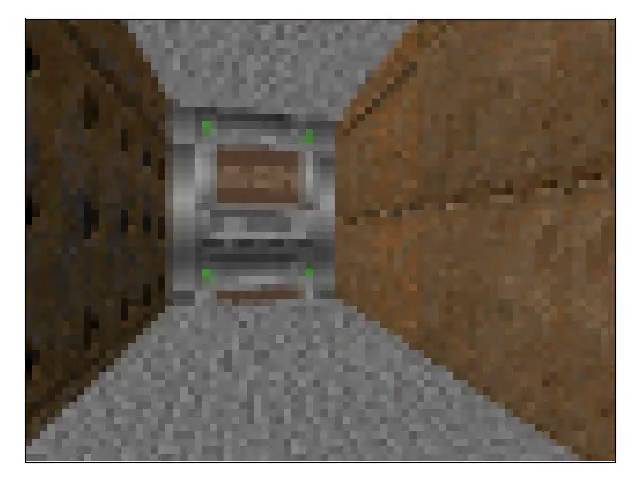}\hspace{-0.2cm}
    \includegraphics[width=0.18\linewidth]{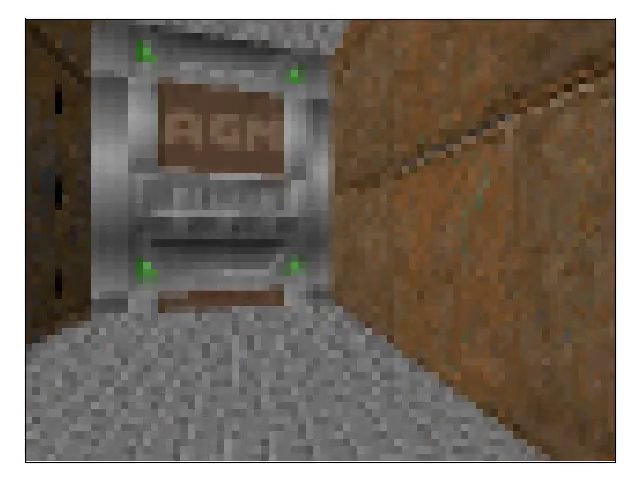}\hspace{-0.2cm}
    \includegraphics[width=0.18\linewidth]{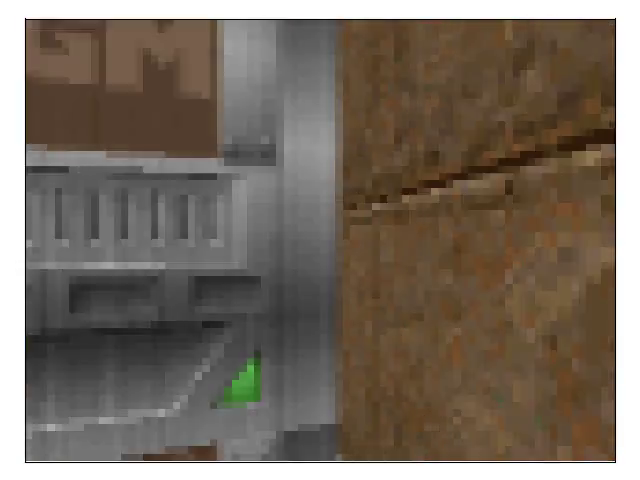}
    \caption{Example trajectories in VizDoom to two goals.}
        \label{fig:doom}
\end{figure}

\begin{figure}[h]
    \centering
    \includegraphics[width=0.75\textwidth]{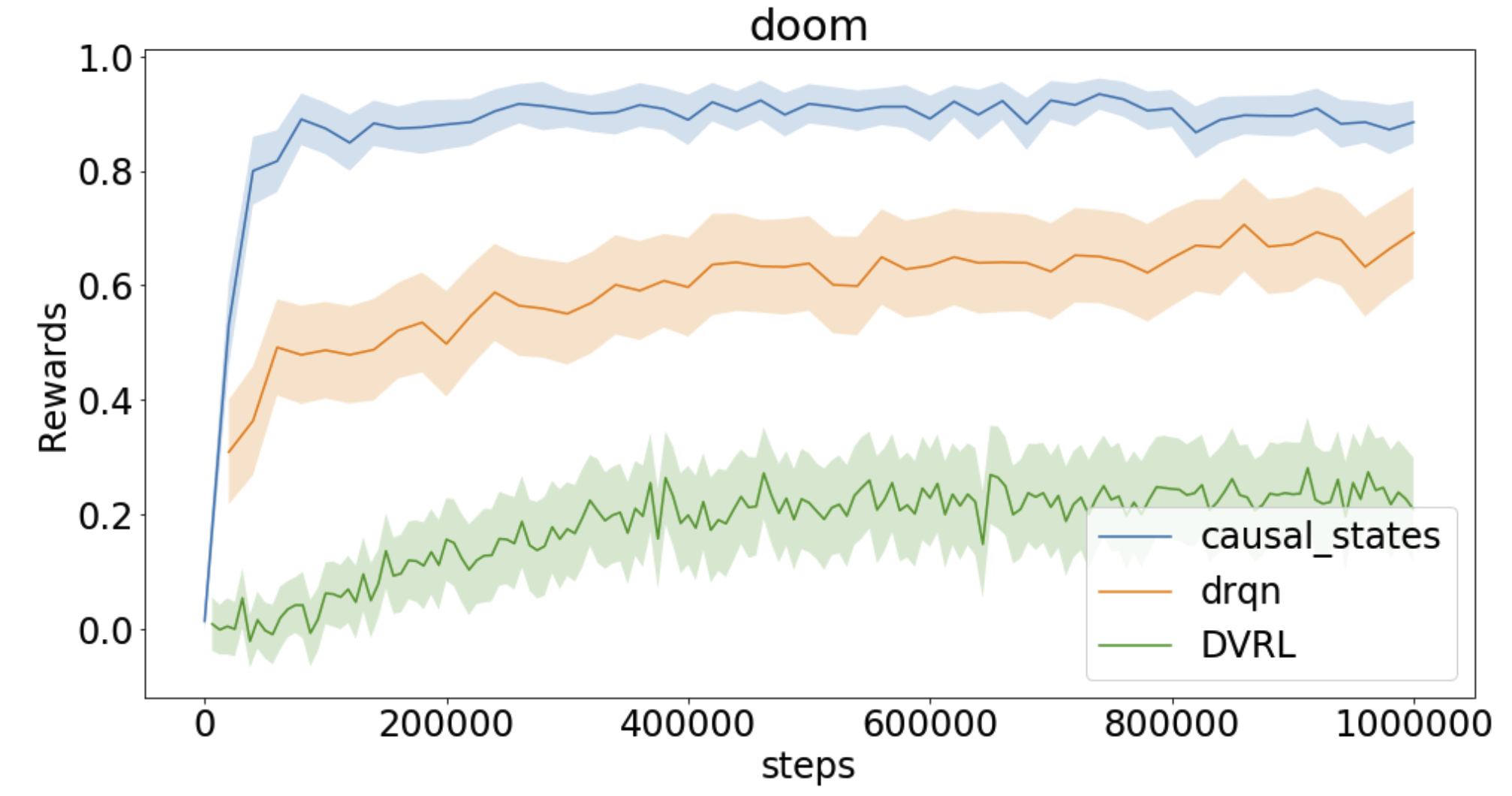}
    \caption{Doom T-Maze POMDP: Averaged over 100 runs with different random seeds with one standard error shaded. Y-axis is mean reward per step. }
    \label{fig:doom_results}
\end{figure}

We now extend our results to continuous state space environments, where causal states are potentially infinite. 
We modify the T-maze VizDoom ~\citep{Kempka2016ViZDoom} environment 
of \cite{DaneSCorneil2018} to make it partially observable. 
We randomize the goal location between the two ending corners 
and signal its location with a stochastic signal in the observation space. 
The agent must remember where the goal is in order to navigate to it. 
We convey the signal to the agent through a fourth channel (after RGB) 
that intermittently contains information about where the goal is. 
The frequency at which the information is displayed 
is a tunable factor $f=2$ for these experiments.
Example trajectories to different goals 
that the learned agent takes are shown in \cref{fig:doom}. We compare with Rainbow DQN~\citep{rainbow} modified with an LSTM to become Rainbow DRQN~\citep{hausknecht2015deep}, and Deep Variational Reinforcement Learning (DVRL)~\citep{igl2018deep}, another gradient-based method for learning belief states.
\cref{fig:doom_results} shows the speedup in learning 
from explicitly learning to cluster sequences of observations into causal states. In \cite{igl2018deep} DVRL was originally evaluated on Atari, and we adapted the implementation to VizDoom but found the method to transfer poorly. Rainbow DRQN performs better, but does not handle the required memory perfectly. Our method, causal states, quickly learns a latent representation maximally predictive of future states and successfully learns a good policy.

\subsection{Flickering Atari}
\label{sec:atari}

\begin{figure}[h]
    \centering
    \includegraphics[height=0.2\textwidth]{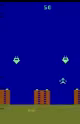} \hspace{1pt}
    \includegraphics[height=0.2\textwidth]{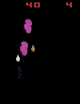} \hspace{1pt}
    \includegraphics[height=0.2\textwidth]{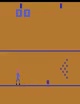} \hspace{1pt}
    \includegraphics[height=0.2\textwidth]{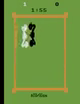} \hspace{1pt}
    \includegraphics[height=0.2\textwidth]{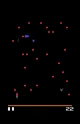} \\ \vspace{5pt}
    \includegraphics[height=0.2\textwidth]{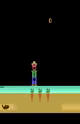} \hspace{1pt}
    \includegraphics[height=0.2\textwidth]{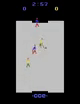} \hspace{1pt}
    \includegraphics[height=0.2\textwidth]{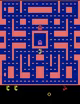} \hspace{1pt}
    \includegraphics[height=0.2\textwidth]{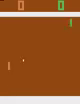} \hspace{1pt}
    \includegraphics[height=0.2\textwidth]{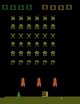}
    \caption{Pixel observations from the 10 Atari games evaluated, in alphabetical order from left to right and top to bottom: \texttt{Air Raid}, \texttt{Asteroids}, \texttt{Bowling}, \texttt{Boxing}, \texttt{Centipede}, \texttt{Gopher}, \texttt{Ice Hockey}, \texttt{Ms. Pacman}, \texttt{Pong}, and \texttt{Space Invaders}.}
    \label{fig:atari_obs}
\end{figure}

We also evaluate our model on a suite of tasks from Atari in OpenAI Gym
\citep{brockman2016openai}, another setting where causal states are less intuitive. Example pixel observations from the tasks evaluated can be found in \cref{fig:atari_obs}.
We use the same preprocessing of observations in \cite{mnih-atari-2013}, except each frame has a $p=0.5$ chance of being blank, introduced as a POMDP version of the environment by \citet{hausknecht2015deep}. 
The environment is now partially observable as the agent must learn 
to retain information when the current observation is blank from the hidden state of the RNN. 
Our algorithm learns an approximate causal state representation suitable
for learning successful policies, achieving a higher final score in most environments evaluated (\cref{tab:atari}).
We again see a decrease in performance with Rainbow DRQN and DVRL, due to instability in training, with better performance exhibited by Rainbow DRQN.

\begin{table}[h]
\vspace{5px} 
    \centering
    \begin{tabular}{l|lll}
    \toprule
    Game & Causal States & DRQN & DVRL \\
    \midrule
     Air Raid & $ \mathbf{950 \pm 271}$ & $518 \pm 231$ & $\mathbf{748 \pm 156}$ \\
     Asteroids & $\mathbf{1129 \pm 345}$ & $\mathbf{929 \pm 285}$ & $349 \pm 54$ \\
     Bowling & $\mathbf{ 34 \pm 8}$ & $\mathbf{29 \pm 0}$ & $23 \pm 1$ \\
     Boxing & $ 4 \pm 4$ & $0 \pm 2$ & $\mathbf{16 \pm 3}$ \\
    Centipede & $\mathbf{4586 \pm 763}$ & $3127 \pm 71$ & $1157 \pm 130$ \\
    Gopher & $ \mathbf{783 \pm 151}$ & $620 \pm 129$ & $255 \pm 129$ \\
    Ice Hockey & $ \mathbf{-3 \pm 1}$ & $\mathbf{-5 \pm 1}$ & $-11 \pm 0$ \\
    Ms. Pacman & $ 671 \pm 36$ & $\mathbf{849 \pm 60}$ & $181 \pm 45$ \\
    Pong & $\mathbf{ -2 \pm 6}$ & $\mathbf{-7 \pm 7}$ & $-20 \pm 0$\\
    Space Invaders & $\mathbf{ 354 \pm 67}$ & $\mathbf{381 \pm 14}$ & $68 \pm 9$\\
     \bottomrule
     \end{tabular}
     \caption{Mean results and $95\%$ confidence intervals across 10 seeds for Atari games evaluated at 2.5M steps.}
     \label{tab:atari}
\end{table}

\section{Related Literature}
In this section we explore other works that have drawn similar parallels across causal states and partial observability methods, as well as related works broaching bisimulation and causal inference, and their corresponding algorithms. We end with an exploration of similar works using knowledge distillation methods.

The relationship between PSRs and causal states has been previously suggested in the computational mechanics literature \citep{shalizi2001computational,shalizi2004blind,barnett2015computational} but not fully explored as an objective for partial observability settings in reinforcement learning. \citet{hundt2006representing} also derive parallels between PSRs, POMDPs, and automata through the construction of equivalence classes that groups states with common action conditional future observations, but do not extend this to the causal literature. Causal states, and the related information-theoretic notion of complexity, minimality, and sufficiency are used to derive task-agnostic policies with an intrinsic exploration-exploitation trade off in  \citet{still2009information,still2012information}. More similar to our latent discrete states with continuous observables, \citet{goerg2013mixed} model spatio-temporal processes as being generated by a finite set of discrete causal states in the form of light-cones. Our additional contribution beyond these works is proving the optimality of the causal state representation for POMDPs, providing additional lower bounds on the optimal state-action value function, and providing a gradient-based algorithm for learning causal state representations.

Now focusing on methods that approach  the partial observability problem in reinforcement learning, \citet{boots2013hilbert} learn PSRs in Reproducing Kernel Hilbert Space, extending the approach to continuous, potentially infinite, action-observation processes. However, their method requires maintaining a state vector with length proportional to the size of the training dataset, which would be untenable in deep RL settings with millions of data points. 
AIXI~\citep{veness2011aixi} is another Bayesian optimality approach for general reinforcement learning in unknown environments, but also relies on maximizing rewards for a finite $m$ steps into the future, as opposed to causal states which (theoretically) optimize for predictive performance into the infinite future.
Other methods for learning deep representations for reinforcement learning POMDPs have been proposed, starting with adding recurrency to DQN~\citep{hausknecht2015deep} to integrate the history in the estimation of the Q-value as opposed to using only the current observation. However, this method stops short of ensuring sufficiency for next step prediction as it learns a task specific representation.
\citet{igl2018deep,hofmann2018pomdp} use deep variational methods to learn a probability distribution over states, i.e., belief states, and use the belief states for policy optimization with actor-critic~\citep{igl2018deep} and planning~\citep{hofmann2018pomdp}.
\citet{guo2018npbr} also use neural methods to learn belief states with next-step prediction,
\citet{hefny2015supervised,hefny2018recurrent} learn PSRs with RNNs and spectral methods and use policy gradient with an alternating optimization method for the policy and the state representation to handle continuous action spaces. Finally, \cite{Jiang2016psrs} propose a gradient-based method for learning a spectral decomposition of the systems dynamic matrix to extract PSRs. 
None of these methods explore the connection to causal states, bisimulation, and compression via a discrete representation, and therefore give no guarantee that the method converges to an optimal solution, or bounds on closeness to that solution. 
\cite{DaneSCorneil2018} do learn discrete representations but not in partially observable environments and with no link to PSRs. Instead, they propose the discrete representation solely for using tabular Q-learning with prioritized sweeping.

Various forms of state abstractions have been defined in Markov decision processes (MDPs) to group states into clusters whilst preserving some property (e.g.\ the optimal value, or all values, or all action values from each state)~\citep{larsen1989bisim,Givan2003EquivalenceNA,li2006stateabs}. The strictest form, which generally preserves the most properties, is \textit{bisimulation}~\citep{larsen1989bisim}. Bisimulation only groups states that are indistinguishable w.r.t.\ reward sequences output given any action sequence tested. A related concept is bisimulation metrics~\citep{ferns2014bisim_metrics}, which measure how ``behaviorally similar'' states are. \citet{ferns2011contbisim} defines the bisimulation metric with respect to continuous MDPs, and propose a Monte Carlo algorithm for learning it using an exact computation of the Wasserstein distance between empirically measured transition distributions. However, this method does not scale well to large state spaces. \citet{taylor2009bounding} relate MDP homomorphisms to lax probabilistic bisimulation, and define a lax bisimulation metric. They then compute a value bound based on this metric for MDP homomorphisms, where approximately equivalent state-action pairs are aggregated. \citet{castro20bisimulation} propose an algorithm for computing \textit{on-policy} bisimulation metrics, but with a focus only on deterministic settings and the policy evaluation problem. \citet{zhang2020dbc} propose a method for learning bisimulation metrics to learn a coarsest bisimulation partition in the control setting, while
\citet{castro2009equivalence}  showed the link between trajectory equivalence  in the partial observability setting and bisimulation.
We believe our work is the first to apply bisimulation-based bounds to learned representations in the partially observable  setting and prove a link to causal states. Similarly, \citet{zhang2020invariant} show the connection between bisimulation and causal feature sets, and utilize the requirements of interventions over all variables in the multi-task reinforcement learning setting with full observability.

The idea of extracting the implicit knowledge of a neural network \citep{towell1993extracting,fu1994rule,lu1996effective,hailesilassie2016rule} is not novel and is rooted in early attempts to merge traditional rule-based methods with machine learning. The most recent examples \citep{weiss2017extracting,wang2017empirical,wang2018comparison} are focused on the ability of character level RNNs to implicitly learn grammars. The only application of these ideas to POMDPs that we are aware of is in \citet{koul2018learning}, where deep recurrent policies \citep{hausknecht2015deep} are quantized into Moore machines. The main difference between our approaches is that we reduce models of the environment, not of the policy, to $\epsilon$-machines.

\section{Discussion}
We conclude with a recap of our contributions, a nuanced analysis of our approach and results, and a discussion of future work.

In this paper, we proposed a self-supervised method 
for learning a minimal sufficient statistic for next step prediction in POMDPs
and articulated its connection to the causal states of a controlled process, as well as the equivalence of causal states and bisimulation. This connection leads us to a new bound on the optimal value function in the MDP generated by the learned causal states representation. We further show connections between causal states and causal feature sets from the causal inference literature, allowing us to borrow explicit requirements about the diversity of data through the language of \textit{interventions}.

Our experiments demonstrate the practical utility of this representation,
both with value iteration for control and exhaustive planning,
in solving infinite-memory POMDP environments (\cref{sec:grid}),
as well as $k$-order MDP environments with high dimensional observations (\cref{sec:doom}, \cref{sec:atari}),
matching the performance achieved with ground truth states. Further, we show that the method extends to settings with continuous state-action spaces with an infinite number of causal states, where we must relax the discrete assumption and only learn a continuous representation, also supported by empirical results in \cref{sec:doom}, \cref{sec:atari}. However, we stress the importance of the discrete causal state results as empirical confirmation that our method is able to reconstruct the correct latent states, and that they are informationally equivalent to the causal states.

Causal states have been mostly studied from the discrete perspective, but in continuous state settings there are potentially an infinite number -- hence the relaxation to a continuous representation in rich observation settings. We also find that discrete optimization is often difficult to tune empirically, and found it to be unstable in rich observation settings. Recent progress in discrete world models~\citep{hafner2020planetv2} may offer a promising approach to a successful empirical implementation for more complex problems. The benefit of the discrete representation is its interpretability and ability to use graph-based planning methods~\citep{zhang2018composable,yang2020plan2vec}. Nevertheless,  our results in the continuous setting are still  promising, in that they empirically outperform other POMDP approaches, while maintaining theoretical guarantees. 

However, we are aware that the empirical choices made to implement the gradient-based method for learning causal states induce some limitations. In this work we chose a two-step method of learning a continuous representation and using knowledge distillation to discretize the continuous causal states. We found this method to be more stable than end-to-end discretization methods like VQ-VAE or straight-through estimators. 

The trade-off is that with a two-step procedure, the performance of the discretized causal states uniquely depends on the performance of the continuous representation. Hence, even in situations where the environment has a finite number of states, we are currently not using this information to improve the robustness or learning speed of the continuous representation. While learning an end-to-end model, we could use the discrete outputs as input for state counting policies or other exploration strategies, potentially reducing the number of samples necessary to learn a model covering all the possible state transitions. Analogously, while learning a reward maximizing policy, discrete causal states could be used in combination with path finding algorithms to efficiently plan more effective exploration. 

To conclude, we think causal states can further be exploited to solve problems other than control by offering a practically and theoretically sound bridge between deep learning, hidden Markov and causal inference representations of dynamical systems.

\acks{T. Furlanello and L. Itti were supported by the National Science Foundation (grant number CCF-1317433), C-BRIC (one of six centers in JUMP, a Semiconductor Research Corporation (SRC) program sponsored by DARPA), and the Intel Corporation. A. Anandkumar is supported in part by Bren endowed chair, Darpa PAI, Raytheon, and Microsoft, Google and Adobe faculty fellowships. K. Azizzadenesheli is supported in part by NSF Career Award CCF-1254106 and AFOSR YIP FA9550-15-1-0221, work done while he was visiting Caltech.  The authors affirm that the views expressed herein are solely their own, and do not represent the views of the United States government or any agency thereof. A. Zhang would like to thank Pablo Samuel Castro, Marc Bellemare, and Doina Precup for helpful discussions and feedback.}

\newpage

\appendix
\section{Proof of \cref{thm:causal_states_bisimilar}}
\label{app:causal_states_bisimilar}
\causalstatesbisimilar*
\begin{proof}
First, we show that if two observation sequences $(o^1_{\leq t},a^1_{<t}),(o^2_{\leq t},a^2_{<t})$ are in the same causal state $\sigma_i$, they are trajectory equivalent. This is evident from \cref{def:causal_states}, where we have
$$\mathbb{P}(O_{>t},A_{\geq t}|o^1_{\leq t},a^1_{<t}) = \mathbb{P}(O_{>t},A_{\geq t}|o^2_{\leq t},a^2_{<t}),$$ therefore conditioned on any action sequence $a_{\geq t}\in A_{\geq t}$ we have $$\mathbb{P}(O_{>t},A_{\geq t}|o^1_{\leq t},a^1_{<t},a_{\geq t}) = \mathbb{P}(O_{>t},A_{\geq t}|o^2_{\leq t},a^2_{<t},a_{\geq t}).$$ 
Denoting the causal state mapping as $\Psi$ and remembering that reward is now part of the observation, gives us $$\mathbb{P}(\alpha|\Psi(s_1),a_{\geq t}) = \mathbb{P}(\alpha|\Psi(s_2),a_{\geq t}),$$
where $s_1:=o^1_{\leq t},a^1_{<t}$ and $s_2:=o^2_{\leq t},a^2_{<t}$. Remember that $\alpha$ is a finite reward-state trajectory and therefore from this equivalence we can pull out, 
\begin{equation*}
\begin{split}
    R(\Psi(s_1),a)&=R(\Psi(s_2),a) \\
    \mathbb{P}(s'|\Psi(s_1),a_{\geq t}) &= \mathbb{P}(s'|\Psi(s_2),a_{\geq t}).
\end{split}
\end{equation*}
where $s'\in\Psi(S)$. 
\end{proof}

\section{Proof of \cref{thm:bisim_bound}}
\label{app:bisim_bound}
\bisimbound*
\begin{proof}
We prove this with the triangle inequality in three steps.
First, we need to show that in the $L_D$-$L_R$-Lipschitz causal state MDP with metric $d$ on the state space the bisimulation metric is bounded,
$$\Tilde{d}(\phi(o^1_{\leq t}),\phi(o^2_{\leq t}))\leq \frac{(1-\gamma)L_R}{1-\gamma L_D}d\big(\phi(o^1_{\leq t}),  \phi(o^2_{\leq t})\big).$$
We do so with a proof by induction for a sequence of psuedometrics $d_n$ that converge to $\Tilde{d}$, starting with $d_0(\phi(o^1_{\leq t}),\phi(o^2_{\leq t}))=0$.

\begin{equation*}
\begin{split}
d_{n+1}(\phi(o^1_{\leq t}),\phi(o^2_{\leq t}))&=\max_{a\in A}\bigg((1-\gamma)|r_{\phi(o^1_{\leq t})}^a-r^a_{\phi(o^2_{\leq t})}|+\gamma W_{d_n} (P_{\phi(o^1_{\leq t})}^a,P_{\phi(o^2_{\leq t})}^a)\bigg) \qquad\text{from \cref{def:bisim_metric}} \\
&= \max_{a\in A}\bigg((1-\gamma)\frac{|r_{\phi(o^1_{\leq t})}^a-r^a_{\phi(o^2_{\leq t})}|}{d(\phi(o^1_{\leq t}),  \phi(o^2_{\leq t}))}+\gamma \frac{W_{d_n} (P_{\phi(o^1_{\leq t})}^a,P_{\phi(o^2_{\leq t})}^a)}{d(\phi(o^1_{\leq t}),  \phi(o^2_{\leq t})}\bigg) d\big(\phi(o^1_{\leq t}),  \phi(o^2_{\leq t}))\big) \\ 
&= \bigg((1-\gamma)\max_{a\in A}\frac{|r_{\phi(o^1_{\leq t})}^a-r^a_{\phi(o^2_{\leq t})}|}{d(\phi(o^1_{\leq t}),  \phi(o^2_{\leq t}))}+\gamma \max_{a\in A}\frac{W_{d_n} (P_{\phi(o^1_{\leq t})}^a,P_{\phi(o^2_{\leq t})}^a)}{d(\phi(o^1_{\leq t}),  \phi(o^2_{\leq t}))}\bigg) d\big(\phi(o^1_{\leq t}),  \phi(o^2_{\leq t})\big) \\
&\leq (1-\gamma)L_R + \gamma \max_{a\in A}\frac{W_{d_n} (P_{\phi(o^1_{\leq t})}^a,P_{\phi(o^2_{\leq t})}^a)}{d(\phi(o^1_{\leq t}),  \phi(o^2_{\leq t}))} d\big(\phi(o^1_{\leq t}),  \phi(o^2_{\leq t})\big) \\
&= (1-\gamma)L_R + \gamma \frac{d_{n}(\phi(o^1_{\leq t}),\phi(o^2_{\leq t}))}{d\big(\phi(o^1_{\leq t}),  \phi(o^2_{\leq t})\big)} \max_{a\in A}\frac{W_{\Tilde{d}} (P_{\phi(o^1_{\leq t})}^a,P_{\phi(o^2_{\leq t})}^a)}{d(\phi(o^1_{\leq t}), \phi(o^2_{\leq t}))} d(\phi(o^1_{\leq t}),  \phi(o^2_{\leq t})) \\
& \hspace{250pt} \text{property of Wasserstein} \\
&\leq (1-\gamma)L_R + \gamma d_{n}(\phi(o^1_{\leq t}),\phi(o^2_{\leq t})) d(\phi(o^1_{\leq t}),  \phi(o^2_{\leq t})) L_D
\end{split}
\end{equation*}
Solving for $n\mapsto\infty$ gives us,
\begin{equation*}
\begin{split}
\tilde{d}(\phi(o^1_{\leq t}),\phi(o^2_{\leq t}))&\leq (1-\gamma)L_R d(\phi(o^1_{\leq t}),  \phi(o^2_{\leq t}))\sum_{i=0}^\infty(\gamma L_D)^i \\
&=\frac{(1-\gamma)L_R}{1-\gamma L_D}d\big(\phi(o^1_{\leq t}),  \phi(o^2_{\leq t})\big).
\end{split}
\end{equation*}

Next, we need to show that if we join the HOMDP and causal state MDP into a single MDP, we can bound the bisimulation metric from a state in one, $o_{\leq t}$, to its corresponding state in the other, $\phi(o_{\leq t})$,
$$\Tilde{d}(o_{\leq t},\phi(o_{\leq t}))\leq \mathcal{L}_r^\infty + \gamma \mathcal{L}_d^\infty\frac{L_R}{1-\gamma L_D}.$$

\begin{equation*}
\begin{split}
    \Tilde{d}(o_{\leq t},\phi(o_{\leq t}))&=\max_{a\in A}\big((1-\gamma)|r^a_{o_{\leq t})}-\bar{r}^a_{\phi(o_{\leq t})}| + \gamma W_{\Tilde{d}}(P(\cdot | o_{\leq t}, a), \bar{P}(\cdot  | \phi(o_{\leq t}), a) \big) \\
    &\leq  (1-\gamma)\mathcal{L}_r^\infty  + \gamma \max_{a\in A} W_{\Tilde{d}}\big(P(\cdot | o_{\leq t}, a), \bar{P}(\cdot  | \phi(o_{\leq t}), a) \big) \\
    &\leq  (1-\gamma)\mathcal{L}_r^\infty  + \gamma \max_{a\in A} 
    \big(W_{\Tilde{d}}\big(P(\cdot | o_{\leq t}, a), \phi P(\cdot  | o_{\leq t}, a) \big) + W_{\Tilde{d}}\big(\phi P(\cdot | o_{\leq t}, a), \bar{P}(\cdot  | \phi(o_{\leq t}), a) \big) \\
    &\leq (1-\gamma)\mathcal{L}_r^\infty  + \gamma \frac{(1-\gamma)L_R}{1-\gamma L_D}\mathcal{L}_d^\infty + \gamma \max_{a\in A}W_{\Tilde{d}}\big(\phi P(\cdot | o_{\leq t}, a), \bar{P}(\cdot  | \phi(o_{\leq t}), a) \big) \\
    &\leq (1-\gamma)\mathcal{L}_r^\infty  + \gamma \frac{(1-\gamma)L_R}{1-\gamma L_D}\mathcal{L}_d^\infty + \gamma \Tilde{d}(o_{\leq t + 1},\phi(o_{\leq t + 1})) \\
    &= \mathcal{L}_r^\infty + \gamma \mathcal{L}_d^\infty\frac{L_R}{1-\gamma L_D} \qquad \text{By recursion.}
\end{split}
\end{equation*}

Now, we can plug these into the triangle inequality,
\begin{equation*}
\begin{split}
\Tilde{d}(o^1_{\leq t},o^2_{\leq t})
&\leq \Tilde{d}(\phi(o^1_{\leq t}),\phi(o^2_{\leq t})) + \Tilde{d}(\phi(o^1_{\leq t}),o^2_{\leq t}) + \Tilde{d}(o^1_{\leq t},\phi(o^2_{\leq t})) \\
&\leq \Tilde{d}(\phi(o^1_{\leq t}),\phi(o^2_{\leq t})) + 2\bigg(\mathcal{L}_r^\infty + \gamma \mathcal{L}_d^\infty\frac{L_R}{1-\gamma L_D}\bigg) \\
&\leq \frac{(1-\gamma)L_R}{1-\gamma L_D}d(\phi(o^1_{\leq t}),  \phi(o^2_{\leq t})) + 2\bigg( \mathcal{L}_r^\infty + \gamma \mathcal{L}_d^\infty\frac{L_R}{1-\gamma L_D}\bigg).
\end{split}
\end{equation*}
\end{proof}

\section{Proof of \cref{thm:valuefn_bound}}
\label{app:valuefn_bound}
\valuefnbound*
\begin{proof}
\begin{align*}
&\sup_{o_{\leq t}\in\mathcal{O}_{\leq t},a_t\in\mathcal{A}}|Q^\pi(o_{\leq t},a_t) - \bar{Q}^\pi (\phi(o_{\leq t}),a_t)|\\
&\qquad\leq \sup_{o_{\leq t}\in\mathcal{O}_{\leq t},a_t\in\mathcal{A}}|R(\phi(o_{\leq t}),a,\phi(o_{\leq t+1})) - r| \\
& \hspace{100pt} + \gamma \sup_{o_{\leq t}\in\mathcal{O}_{\leq t},a_t\in\mathcal{A}}|\mathbb{E}_{o_{\leq t+1}\sim P(\cdot|o_{\leq t},a_t)}V^\pi(o_{\leq t+1}) - \mathbb{E}_{\hat{s}_{t+1}\sim f(\cdot|\phi(o_{\leq t}),a_t)}\bar{V}^\pi(\hat{s}_{t+1})| \\
&\qquad= \mathcal{L}^\infty_r + \gamma \sup_{o_{\leq t}\in\mathcal{O}_{\leq t},a_t\in\mathcal{A}}\big|\mathbb{E}_{o_{\leq t+1}\sim P(\cdot|o_{\leq t},a_t)}[V^\pi(o_{\leq t+1}) - \bar{V}^\pi(\phi(o_{\leq t+1}))] \\
&\qquad \phantom{= \mathcal{L}^\infty_r~} + \mathbb{E}_{\substack{o_{\leq t+1}\sim P(\cdot|o_{\leq t},a_t) \\ \hat{s}_{t+1}\sim f(\cdot|\phi(o_{\leq t}),a_t)}}[\bar{V}^\pi(\phi(o_{\leq t+1})) - \bar{V}^\pi(\hat{s}_{t+1})]\big| \\
&\qquad \leq \mathcal{L}^\infty_r + \gamma \sup_{o_{\leq t}\in\mathcal{O}_{\leq t},a_t\in\mathcal{A}}\big|\mathbb{E}_{o_{\leq t+1}\sim P(\cdot|o_{\leq t},a_t)}[V^\pi(o_{\leq t+1}) - \bar{V}^\pi(\phi(o_{\leq t+1}))]\big|\\
&\qquad \phantom{\leq \mathcal{L}^\infty_r~} + \gamma \sup_{o_{\leq t}\in\mathcal{O}_{\leq t},a_t\in\mathcal{A}} \big| \mathbb{E}_{\substack{o_{\leq t+1}\sim P(\cdot|o_{\leq t},a_t) \\ \hat{s}_{t+1}\sim f(\cdot|\phi(o_{\leq t}),a_t)}}[\bar{V}^\pi(\phi(o_{\leq t+1})) - \bar{V}^\pi(\hat{s}_{t+1})]\big| \\
&\qquad\leq \mathcal{L}^\infty_r + \gamma \sup_{o_{\leq t}\in\mathcal{O}_{\leq t},a_t\in\mathcal{A}}\big|\mathbb{E}_{o_{\leq t+1}\sim P(\cdot|o_{\leq t},a_t)}[V^\pi(o_{\leq t+1}) - \bar{V}^\pi(\phi(o_{\leq t+1}))]\big|\\
&\qquad \phantom{\leq \mathcal{L}^\infty_r~}+ \gamma L_{V^*} \sup_{o_{\leq t}\in\mathcal{O}_{\leq t},a_t\in\mathcal{A}} W_d(\phi(P(\cdot|o_{\leq t},a_t)), f(\cdot|\phi(o_{\leq t}),a_t)) \\
&\qquad = \mathcal{L}^\infty_r + \gamma \sup_{o_{\leq t}\in\mathcal{O}_{\leq t},a_t\in\mathcal{A}}\big|\mathbb{E}_{o_{\leq t+1}\sim P(\cdot|o_{\leq t},a_t)}[V^\pi(o_{\leq t+1}) - \bar{V}^\pi(\phi(o_{\leq t+1}))]\big| + \gamma L_{V^*} \mathcal{L}^\infty_d\\
&\qquad\leq \mathcal{L}^\infty_r + \gamma \sup_{o_{\leq t}\in\mathcal{O}_{\leq t},a_t\in\mathcal{A}}\mathbb{E}_{o_{\leq t+1}\sim P(\cdot|o_{\leq t},a_t)}\big|[V^\pi(o_{\leq t+1}) - \bar{V}^\pi(\phi(o_{\leq t+1}))]\big| + \gamma L_{V^*} \mathcal{L}^\infty_d\\
&\qquad\leq \mathcal{L}^\infty_r + \gamma \sup_{o_{\leq t}\in\mathcal{O}_{\leq t},a_t\in\mathcal{A}}\big|[V^\pi(o_{\leq t}) - \bar{V}^\pi(\phi(o_{\leq t}))]\big| + \gamma L_{V^*} \mathcal{L}^\infty_d\\
&\qquad\leq \mathcal{L}^\infty_r + \gamma \sup_{o_{\leq t}\in\mathcal{O}_{\leq t},a_t\in\mathcal{A}}\big|[Q^\pi(x_{t-1},a_{t-1}) - \bar{Q}^\pi(\phi(x_{t-1}),a_{t-1})]\big| + \gamma L_{V^*} \mathcal{L}^\infty_d\\
&\qquad = \frac{\mathcal{L}^\infty_r + \gamma L_{V^*}\mathcal{L}^\infty_d}{1-\gamma}
\end{align*}
\end{proof}

\section{Proof of \cref{thm:causalstatesets}}
\causalstatesets*
\begin{proof}
To prove that $\phi_S$ is a bisimulation, we must first show that $r(x) = r(x')$ for any $x, x': \phi_S(x) = \phi_S(x')$.
For this, we note that $\mathbb{E}[R(x)] = \int_{r \in \mathbb{R}} r dp(r|x) = \int_{r \in \mathbb{R}} r dp(r|[x]_S, [x]_{S^C})$ and, because by definition $S^C \subset \textbf{PA}(R)^C$, we have that $R \perp [x]_{S^C}$. Therefore, 
\begin{equation}
    \mathbb{E}[R(x)] = \int_{r \in \mathbb{R}} r dp(r|[x]_S) = \int_{r \in \mathbb{R}} r dp(r|[x']_S) = \mathbb{E}[R(x')].
\end{equation}

To show that $[x]_S$ is a bisimulation, we must also show that for any $x_1, x_2 $ such that $\phi(x_1) = \phi(x_2)$, and for any $e \in \mathcal{E}$, the distribution over next state equivalence classes will be equal for $x_1$ and $x_2$:
$$\sum_{x' \in \phi^{-1}(\bar{X})} P^e_{x_1x'} = \sum_{x' \in \phi^{-1}(\bar{X})} P^e_{x_2x'}.$$  

For this, it suffices to observe that $S$ is closed under taking parents in the causal graph, and that by construction environments only contain interventions on variables outside of the causal set.  Specifically, we observe that the probability of seeing any particular equivalence class $[x']_S$ after state $x$ is only a function of $[x]_S$: 
$P([x']_S | x) = f([x]_S, [x']_S)$. This allows us to define a natural decomposition of the transition function as follows:
    $P(x' | x) = P\bigg ( [x]_S \oplus [x]_{S^C} \bigg| [x']_S \oplus [x']_{S^C} \bigg)$, which by the independent noise assumption gives 
    $P(x'|x) =f([x']_S, [x]_S) P([x']_{S^c}|x).$
We further observe that since the components of $x$ are independent, $\sum_{[x']_{S^C}} P([x']_{S^C}|x) = 1$.
We now return to the property we want to show:
\begin{align*}
    \sum_{x' \in \phi^{-1}(\bar{x})} P^e_{x_1x'} &= \sum_{x' \in \phi^{-1}(\bar{x})} f([x_1]_S, [x']_S) P(x' | x_1 ) \\
    &= f(\phi(x_1), \bar{x})\sum_{[x']_{S^C}} P\bigg([x']_{S^C} \bigg | x_1 \bigg)\\
    &= f(\phi(x_1), \bar{x}) \\
\intertext{and because $\phi(x_1) = \phi(x_2)$, we have}
    &= f(\phi(x_2), \bar{x}) \\
\intertext{for which we can apply the previous chain of equalities backward to obtain}
&= \sum_{x' \in \phi^{-1}(\bar{x})} P^e_{x_2x'}  \\
\end{align*}

\end{proof}

\section{Additional Gridworld Results}
\label{sec:add_gridworld}
\begin{figure}[t]
    \centering
    \includegraphics[width=0.47\textwidth]{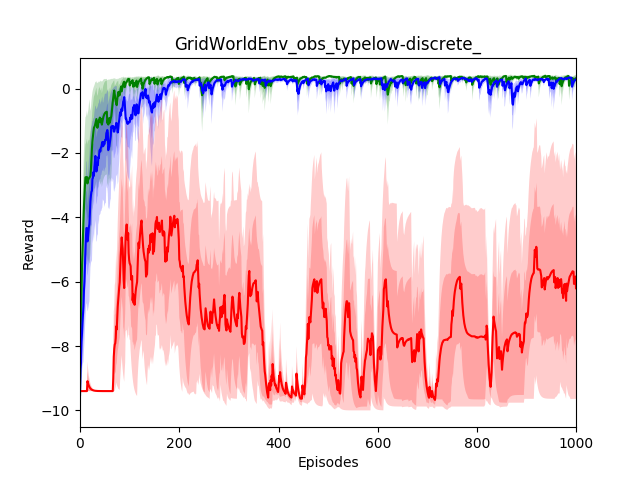}\hspace{-10pt}
    \includegraphics[width=0.47\textwidth]{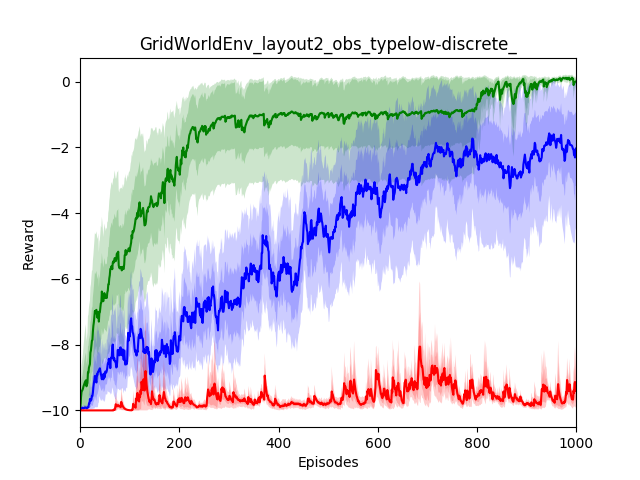}
    \caption{Training curves for DQN policies---discretization with gradient descent and bottleneck networks \textit{(left)} Layout 1 using discrete inputs.\textit{(right)} Layout 2 using discrete inputs.  Averages over 10 runs with different random seeds with two standard deviations shaded. Y-axis is mean reward per step. Green is continuous causal states, Blue is discrete causal states, red is DRQN.}
    \label{fig:disc_grid}
\end{figure}

In \cref{fig:disc_grid} we have additional results using ternary neurons as a gradient-based approach to discretization. We compare the continuous representation, which performs best, to the discretized causal states in blue, and DRQN in red. We find a drop in performance when performing discretization, showing that we are not able to empirically retain the required information with this discretization method.

\bibliography{refs}

\begin{thebibliography}{70}
\providecommand{\natexlab}[1]{#1}
\providecommand{\url}[1]{\texttt{#1}}
\expandafter\ifx\csname urlstyle\endcsname\relax
  \providecommand{\doi}[1]{doi: #1}\else
  \providecommand{\doi}{doi: \begingroup \urlstyle{rm}\Url}\fi

\bibitem[Aastrom(1965)]{aastrom1965optimal}
Karl~J Aastrom.
\newblock Optimal control of markov processes with incomplete state
  information.
\newblock \emph{Journal of Mathematical Analysis and Applications}, 10\penalty0
  (1):\penalty0 174--205, 1965.

\bibitem[{Arjovsky} et~al.(2019){Arjovsky}, {Bottou}, {Gulrajani}, and
  {Lopez-Paz}]{arjovsky2019irm}
Martin {Arjovsky}, Leon {Bottou}, Ishaan {Gulrajani}, and David {Lopez-Paz}.
\newblock {Invariant Risk Minimization}.
\newblock \emph{arXiv e-prints}, July 2019.

\bibitem[Azizzadenesheli et~al.(2016)Azizzadenesheli, Lazaric, and
  Anandkumar]{azizzadenesheli2016reinforcement}
Kamyar Azizzadenesheli, Alessandro Lazaric, and Animashree Anandkumar.
\newblock Reinforcement learning in rich-observation {MDPs} using spectral
  methods.
\newblock \emph{arXiv preprint arXiv:1611.03907}, 2016.

\bibitem[Barnett and Crutchfield(2015)]{barnett2015computational}
Nix Barnett and James~P Crutchfield.
\newblock Computational mechanics of input--output processes: Structured
  transformations and the eps-transducer.
\newblock \emph{Journal of Statistical Physics}, 161\penalty0 (2):\penalty0
  404--451, 2015.

\bibitem[Bellemare et~al.(2013)Bellemare, Naddaf, Veness, and
  Bowling]{bellemare13arcade}
Marc~G. Bellemare, Yavar Naddaf, Joel Veness, and Michael Bowling.
\newblock The arcade learning environment: An evaluation platform for general
  agents.
\newblock \emph{J. Artif. Int. Res.}, 47\penalty0 (1):\penalty0 253–279, May
  2013.
\newblock ISSN 1076-9757.

\bibitem[Bertsekas and Castanon(1989)]{bertsekas1989bounds}
Dimitri Bertsekas and David Castanon.
\newblock Adaptive aggregation for infinite horizon dynamic programming.
\newblock \emph{Automatic Control, IEEE Transactions on}, 34:\penalty0 589 --
  598, 07 1989.
\newblock \doi{10.1109/9.24227}.

\bibitem[Boots et~al.(2013)Boots, Gordon, and Gretton]{boots2013hilbert}
Byron Boots, Geoffrey Gordon, and Arthur Gretton.
\newblock Hilbert space embeddings of predictive state representations.
\newblock \emph{arXiv preprint arXiv:1309.6819}, 2013.

\bibitem[Brockman et~al.(2016)Brockman, Cheung, Pettersson, Schneider,
  Schulman, Tang, and Zaremba]{brockman2016openai}
Greg Brockman, Vicki Cheung, Ludwig Pettersson, Jonas Schneider, John Schulman,
  Jie Tang, and Wojciech Zaremba.
\newblock Openai gym, 2016.
\newblock URL \url{http://arxiv.org/abs/1606.01540}.
\newblock cite arxiv:1606.01540.

\bibitem[Cassandra et~al.(1994)Cassandra, Kaelbling, and
  Littman]{cassandra1994acting}
Anthony~R Cassandra, Leslie~Pack Kaelbling, and Michael~L Littman.
\newblock Acting optimally in partially observable stochastic domains.
\newblock In \emph{AAAI}, volume~94, pages 1023--1028, 1994.

\bibitem[Castro(2020)]{castro20bisimulation}
Pablo~Samuel Castro.
\newblock Scalable methods for computing state similarity in deterministic
  {M}arkov decision processes.
\newblock In \emph{Association for the Advancement of Artificial Intelligence
  (AAAI)}, 2020.

\bibitem[Castro and Precup(2010)]{castro2010using}
Pablo~Samuel Castro and Doina Precup.
\newblock Using bisimulation for policy transfer in {MDPs}.
\newblock In \emph{Twenty-Fourth AAAI Conference on Artificial Intelligence},
  2010.

\bibitem[Castro et~al.(2009)Castro, Panangaden, and
  Precup]{castro2009equivalence}
Pablo~Samuel Castro, Prakash Panangaden, and Doina Precup.
\newblock Equivalence relations in fully and partially observable markov
  decision processes.
\newblock In \emph{Proceedings of the 21st International Jont Conference on
  Artifical Intelligence}, IJCAI'09, pages 1653--1658, San Francisco, CA, USA,
  2009. Morgan Kaufmann Publishers Inc.
\newblock URL \url{http://dl.acm.org/citation.cfm?id=1661445.1661711}.

\bibitem[Chung et~al.(2014)Chung, Gulcehre, Cho, and
  Bengio]{chung2014empirical}
Junyoung Chung, Caglar Gulcehre, KyungHyun Cho, and Yoshua Bengio.
\newblock Empirical evaluation of gated recurrent neural networks on sequence
  modeling.
\newblock \emph{arXiv preprint arXiv:1412.3555}, 2014.

\bibitem[Corneil et~al.(2018)Corneil, Gerstner, and Brea]{DaneSCorneil2018}
Dane~S. Corneil, Wulfram Gerstner, and Johanni Brea.
\newblock Efficient model-based deep reinforcement learning with variational
  state tabulation.
\newblock In \emph{Proceedings of the 35th International Conference on Machine
  Learning, {ICML} 2018, Stockholmsm{\"{a}}ssan, Stockholm, Sweden, July 10-15,
  2018}, pages 1057--1066, 2018.

\bibitem[Crutchfield and Young(1989)]{crutchfield1989inferring}
James~P Crutchfield and Karl Young.
\newblock Inferring statistical complexity.
\newblock \emph{Physical Review Letters}, 63\penalty0 (2):\penalty0 105, 1989.

\bibitem[Crutchfield et~al.(2010)Crutchfield, Ellison, James, and
  Mahoney]{crutchfield2010synchronization}
James~P Crutchfield, Christopher~J Ellison, Ryan~G James, and John~R Mahoney.
\newblock Synchronization and control in intrinsic and designed computation: An
  information-theoretic analysis of competing models of stochastic computation.
\newblock \emph{Chaos: An Interdisciplinary Journal of Nonlinear Science},
  20\penalty0 (3):\penalty0 037105, 2010.

\bibitem[Eberhardt and Scheines(2007)]{eberhardt2007interventions}
Frederick Eberhardt and Richard Scheines.
\newblock Interventions and causal inference.
\newblock \emph{Philosophy of Science}, 74\penalty0 (5):\penalty0 981--995,
  2007.
\newblock \doi{10.1086/525638}.
\newblock URL \url{https://doi.org/10.1086/525638}.

\bibitem[Ferns et~al.(2004)Ferns, Panangaden, and
  Precup]{ferns2004bisimulation}
Norm Ferns, Prakash Panangaden, and Doina Precup.
\newblock Metrics for finite markov decision processes.
\newblock In \emph{Proceedings of the 20th Conference on Uncertainty in
  Artificial Intelligence}, UAI '04, pages 162--169, Arlington, Virginia,
  United States, 2004. AUAI Press.
\newblock ISBN 0-9749039-0-6.
\newblock URL \url{http://dl.acm.org/citation.cfm?id=1036843.1036863}.

\bibitem[Ferns et~al.(2011)Ferns, Panangaden, and Precup]{ferns2011contbisim}
Norm Ferns, Prakash Panangaden, and Doina Precup.
\newblock Bisimulation metrics for continuous markov decision processes.
\newblock \emph{SIAM J. Comput.}, 40\penalty0 (6):\penalty0 1662--1714,
  December 2011.
\newblock ISSN 0097-5397.
\newblock \doi{10.1137/10080484X}.
\newblock URL \url{http://dx.doi.org/10.1137/10080484X}.

\bibitem[Ferns and Precup(2014)]{ferns2014bisim_metrics}
Norman Ferns and Doina Precup.
\newblock Bisimulation metrics are optimal value functions.
\newblock In \emph{Uncertainty in Artificial Intelligence (UAI)}, pages
  210--219, 2014.

\bibitem[Fu(1994)]{fu1994rule}
LiMin Fu.
\newblock Rule generation from neural networks.
\newblock \emph{IEEE Transactions on Systems, Man, and Cybernetics},
  24\penalty0 (8):\penalty0 1114--1124, 1994.

\bibitem[Gelada et~al.(2019)Gelada, Kumar, Buckman, Nachum, and
  Bellemare]{gelada2019deepmdp}
Carles Gelada, Saurabh Kumar, Jacob Buckman, Ofir Nachum, and Marc~G.
  Bellemare.
\newblock {D}eep{MDP}: Learning continuous latent space models for
  representation learning.
\newblock In Kamalika Chaudhuri and Ruslan Salakhutdinov, editors,
  \emph{Proceedings of the 36th International Conference on Machine Learning},
  volume~97 of \emph{Proceedings of Machine Learning Research}, pages
  2170--2179, Long Beach, California, USA, 09--15 Jun 2019. PMLR.

\bibitem[Givan et~al.(2003)Givan, Dean, and Greig]{Givan2003EquivalenceNA}
Robert Givan, Thomas~L. Dean, and Matthew Greig.
\newblock Equivalence notions and model minimization in markov decision
  processes.
\newblock \emph{Artif. Intell.}, 147:\penalty0 163--223, 2003.

\bibitem[Goerg and Shalizi(2013)]{goerg2013mixed}
Georg Goerg and Cosma Shalizi.
\newblock Mixed licors: A nonparametric algorithm for predictive state
  reconstruction.
\newblock In \emph{Artificial Intelligence and Statistics}, pages 289--297,
  2013.

\bibitem[Guo et~al.(2018)Guo, Azar, Piot, Pires, Pohlen, and
  Munos]{guo2018npbr}
Zhaohan~Daniel Guo, Mohammad~Gheshlaghi Azar, Bilal Piot, Bernardo~A. Pires,
  Toby Pohlen, and R{\'{e}}mi Munos.
\newblock Neural predictive belief representations.
\newblock \emph{CoRR}, abs/1811.06407, 2018.

\bibitem[Hafner et~al.(2020)Hafner, Lillicrap, Norouzi, and
  Ba]{hafner2020planetv2}
Danijar Hafner, Timothy Lillicrap, Mohammad Norouzi, and Jimmy Ba.
\newblock Mastering atari with discrete world models, 2020.

\bibitem[Hailesilassie(2016)]{hailesilassie2016rule}
Tameru Hailesilassie.
\newblock Rule extraction algorithm for deep neural networks: A review.
\newblock \emph{arXiv preprint arXiv:1610.05267}, 2016.

\bibitem[Hausknecht and Stone(2015)]{hausknecht2015deep}
Matthew Hausknecht and Peter Stone.
\newblock Deep recurrent q-learning for partially observable {MDPs}, 2015.
\newblock URL
  \url{https://www.aaai.org/ocs/index.php/FSS/FSS15/paper/view/11673}.

\bibitem[Hefny et~al.(2015)Hefny, Downey, and Gordon]{hefny2015supervised}
Ahmed Hefny, Carlton Downey, and Geoffrey~J Gordon.
\newblock Supervised learning for dynamical system learning.
\newblock In \emph{Advances in neural information processing systems}, pages
  1963--1971, 2015.

\bibitem[Hefny et~al.(2018)Hefny, Marinho, Sun, Srinivasa, and
  Gordon]{hefny2018recurrent}
Ahmed Hefny, Zita Marinho, Wen Sun, Siddhartha Srinivasa, and Geoffrey Gordon.
\newblock Recurrent predictive state policy networks.
\newblock \emph{arXiv preprint arXiv:1803.01489}, 2018.

\bibitem[Hessel et~al.(2017)Hessel, Modayil, van Hasselt, Schaul, Ostrovski,
  Dabney, Horgan, Piot, Azar, and Silver]{rainbow}
Matteo Hessel, Joseph Modayil, Hado van Hasselt, Tom Schaul, Georg Ostrovski,
  Will Dabney, Dan Horgan, Bilal Piot, Mohammad~Gheshlaghi Azar, and David
  Silver.
\newblock Rainbow: Combining improvements in deep reinforcement learning.
\newblock In \emph{AAAI}, 2017.

\bibitem[Hinton(2013)]{hinton2013rnn}
Geoffrey Hinton.
\newblock Recurrent neural networks, lecture 10 of {CSC2535}.
\newblock 2013.

\bibitem[Hinton et~al.(2015)Hinton, Vinyals, and Dean]{hinton2015distilling}
Geoffrey Hinton, Oriol Vinyals, and Jeff Dean.
\newblock Distilling the knowledge in a neural network.
\newblock \emph{arXiv preprint arXiv:1503.02531}, 2015.

\bibitem[Hundt et~al.(2007)Hundt, Panangaden, Pineau, and
  Precup]{hundt2006representing}
Christopher Hundt, Prakash Panangaden, Joelle Pineau, and Doina Precup.
\newblock Representing systems with hidden state.
\newblock In Clayton~T. Morrison and Tim Oates, editors, \emph{Computational
  Approaches to Representation Change during Learning and Development, Papers
  from the 2007 {AAAI} Fall Symposium, Arlington, Virginia, USA, November 9-11,
  2007}, volume {FS-07-03} of \emph{{AAAI} Technical Report}, pages 17--23.
  {AAAI} Press, 2007.
\newblock URL
  \url{https://www.aaai.org/Library/Symposia/Fall/2007/fs07-03-003.php}.

\bibitem[Igl et~al.(2018)Igl, Zintgraf, Le, Wood, and Whiteson]{igl2018deep}
Maximilian Igl, Luisa Zintgraf, Tuan~Anh Le, Frank Wood, and Shimon Whiteson.
\newblock Deep variational reinforcement learning for {POMDPs}.
\newblock \emph{arXiv preprint arXiv:1806.02426}, 2018.

\bibitem[Jiang et~al.(2016)Jiang, Kulesza, and Singh]{Jiang2016psrs}
Nan Jiang, Alex Kulesza, and Satinder Singh.
\newblock Improving predictive state representations via gradient descent.
\newblock In \emph{Proceedings of the Thirtieth AAAI Conference on Artificial
  Intelligence}, AAAI'16, pages 1709--1715. AAAI Press, 2016.
\newblock URL \url{http://dl.acm.org/citation.cfm?id=3016100.3016138}.

\bibitem[Kaelbling et~al.(1998)Kaelbling, Littman, and
  Cassandra]{kaelbling1998planning}
Leslie~Pack Kaelbling, Michael~L Littman, and Anthony~R Cassandra.
\newblock Planning and acting in partially observable stochastic domains.
\newblock \emph{Artificial intelligence}, 101\penalty0 (1-2):\penalty0 99--134,
  1998.

\bibitem[Kempka et~al.(2016)Kempka, Wydmuch, Runc, Toczek, and
  Ja\'skowski]{Kempka2016ViZDoom}
Micha{\l} Kempka, Marek Wydmuch, Grzegorz Runc, Jakub Toczek, and Wojciech
  Ja\'skowski.
\newblock {ViZDoom}: A {D}oom-based {AI} research platform for visual
  reinforcement learning.
\newblock In \emph{IEEE Conference on Computational Intelligence and Games},
  pages 341--348, Santorini, Greece, Sep 2016. IEEE.
\newblock URL \url{http://arxiv.org/abs/1605.02097}.
\newblock The best paper award.

\bibitem[Koul et~al.(2018)Koul, Greydanus, and Fern]{koul2018learning}
Anurag Koul, Sam Greydanus, and Alan Fern.
\newblock Learning finite state representations of recurrent policy networks.
\newblock \emph{arXiv preprint arXiv:1811.12530}, 2018.

\bibitem[Larsen and Skou(1989)]{larsen1989bisim}
Kim~G. Larsen and Arne Skou.
\newblock Bisimulation through probabilistic testing (preliminary report).
\newblock In \emph{Proceedings of the 16th ACM SIGPLAN-SIGACT Symposium on
  Principles of Programming Languages}, POPL ’89, page 344–352, New York,
  NY, USA, 1989. Association for Computing Machinery.
\newblock ISBN 0897912942.
\newblock \doi{10.1145/75277.75307}.
\newblock URL \url{https://doi.org/10.1145/75277.75307}.

\bibitem[Li et~al.(2006)Li, Walsh, and Littman]{li2006stateabs}
Lihong Li, Thomas Walsh, and Michael Littman.
\newblock Towards a unified theory of state abstraction for {MDPs}.
\newblock \emph{Proceedings of the Ninth International Symposium on Artificial
  Intelligence and Mathematics}, 01 2006.

\bibitem[Littman and Sutton(2001)]{littman2002predictive}
Michael~L Littman and Richard~S Sutton.
\newblock Predictive representations of state.
\newblock In \emph{Advances in neural information processing systems}, pages
  1555--1561, 2001.

\bibitem[Lu et~al.(1996)Lu, Setiono, and Liu]{lu1996effective}
Hongjun Lu, Rudy Setiono, and Huan Liu.
\newblock Effective data mining using neural networks.
\newblock \emph{IEEE transactions on knowledge and data engineering},
  8\penalty0 (6):\penalty0 957--961, 1996.

\bibitem[Mnih et~al.(2013)Mnih, Kavukcuoglu, Silver, Graves, Antonoglou,
  Wierstra, and Riedmiller]{mnih-atari-2013}
Volodymyr Mnih, Koray Kavukcuoglu, David Silver, Alex Graves, Ioannis
  Antonoglou, Daan Wierstra, and Martin Riedmiller.
\newblock Playing atari with deep reinforcement learning.
\newblock In \emph{NIPS Deep Learning Workshop}. 2013.

\bibitem[Pearl(2009)]{pearl2009do}
Judea Pearl.
\newblock \emph{Causality: Models, Reasoning and Inference}.
\newblock Cambridge University Press, New York, NY, USA, 2nd edition, 2009.
\newblock ISBN 052189560X, 9780521895606.

\bibitem[Peters et~al.(2016)Peters, B\"uhlmann, and Meinshausen]{peters2016icp}
Jonas Peters, Peter B\"uhlmann, and Nicolai Meinshausen.
\newblock Causal inference using invariant prediction: identification and
  confidence intervals.
\newblock \emph{Journal of the Royal Statistical Society, Series B (with
  discussion)}, 78\penalty0 (5):\penalty0 947--1012, 2016.

\bibitem[Roy(2006)]{Roy06sabounds}
Benjamin~Van Roy.
\newblock Performance loss bounds for approximate value iteration with state
  aggregation.
\newblock \emph{Math. Oper. Res.}, 31\penalty0 (2):\penalty0 234--244, 2006.
\newblock \doi{10.1287/moor.1060.0188}.
\newblock URL \url{https://doi.org/10.1287/moor.1060.0188}.

\bibitem[Roy et~al.(2005)Roy, Gordon, and Thrun]{roy2005compressedsensing}
Nicholas Roy, Geoffrey Gordon, and Sebastian Thrun.
\newblock Finding approximate {POMDP} solutions through belief compression.
\newblock \emph{J. Artif. Int. Res.}, 23\penalty0 (1):\penalty0 1–40, January
  2005.
\newblock ISSN 1076-9757.

\bibitem[Schmidhuber(1990{\natexlab{a}})]{schmidhuber1990rnnpomdp}
J{\"u}rgen Schmidhuber.
\newblock Learning algorithms for networks with internal and external feedback.
\newblock \emph{Proc. of the 1990 Connectionist Models Summer School}, pages
  52--61, 1990{\natexlab{a}}.

\bibitem[Schmidhuber(1990{\natexlab{b}})]{Schmidhuber90anon-line}
Jürgen Schmidhuber.
\newblock An on-line algorithm for dynamic reinforcement learning and planning
  in reactive environments.
\newblock In \emph{In Proc. IEEE/INNS International Joint Conference on Neural
  Networks}, pages 253--258. IEEE Press, 1990{\natexlab{b}}.

\bibitem[Sch\"{o}lkopf et~al.(2012)Sch\"{o}lkopf, Janzing, Peters, Sgouritsa,
  Zhang, and Mooij]{scholkopf2012causal}
Bernhard Sch\"{o}lkopf, Dominik Janzing, Jonas Peters, Eleni Sgouritsa, Kun
  Zhang, and Joris Mooij.
\newblock On causal and anticausal learning.
\newblock In \emph{Proceedings of the 29th International Coference on
  International Conference on Machine Learning}, ICML’12, page 459–466,
  Madison, WI, USA, 2012. Omnipress.
\newblock ISBN 9781450312851.

\bibitem[Schölkopf(2019)]{scholkopf2019causalityml}
Bernhard Schölkopf.
\newblock Causality for machine learning.
\newblock \emph{CoRR}, abs/1911.10500, 2019.

\bibitem[Shalizi and Crutchfield(2001)]{shalizi2001computational}
Cosma~Rohilla Shalizi and James~P Crutchfield.
\newblock Computational mechanics: Pattern and prediction, structure and
  simplicity.
\newblock \emph{Journal of statistical physics}, 104\penalty0 (3-4):\penalty0
  817--879, 2001.

\bibitem[Shalizi and Shalizi(2004)]{shalizi2004blind}
Cosma~Rohilla Shalizi and Kristina~Lisa Shalizi.
\newblock Blind construction of optimal nonlinear recursive predictors for
  discrete sequences.
\newblock In \emph{Proceedings of the 20th conference on Uncertainty in
  artificial intelligence}, pages 504--511. AUAI Press, 2004.

\bibitem[Singh et~al.(2004)Singh, James, and Rudary]{singh2004predictive}
Satinder Singh, Michael~R James, and Matthew~R Rudary.
\newblock Predictive state representations: A new theory for modeling dynamical
  systems.
\newblock In \emph{Proceedings of the 20th conference on Uncertainty in
  artificial intelligence}, pages 512--519. AUAI Press, 2004.

\bibitem[Still(2009)]{still2009information}
Susanne Still.
\newblock Information-theoretic approach to interactive learning.
\newblock \emph{EPL (Europhysics Letters)}, 85\penalty0 (2):\penalty0 28005,
  2009.

\bibitem[Still and Precup(2012)]{still2012information}
Susanne Still and Doina Precup.
\newblock An information-theoretic approach to curiosity-driven reinforcement
  learning.
\newblock \emph{Theory in Biosciences}, 131\penalty0 (3):\penalty0 139--148,
  2012.

\bibitem[Strelioff and Crutchfield(2014)]{strelioff2014bayesian}
Christopher~C Strelioff and James~P Crutchfield.
\newblock Bayesian structural inference for hidden processes.
\newblock \emph{Physical Review E}, 89\penalty0 (4):\penalty0 042119, 2014.

\bibitem[Taylor et~al.(2009)Taylor, Precup, and Panagaden]{taylor2009bounding}
Jonathan Taylor, Doina Precup, and Prakash Panagaden.
\newblock Bounding performance loss in approximate {MDP} homomorphisms.
\newblock In \emph{Neural Information Processing (NeurIPS)}, pages 1649--1656,
  2009.
\newblock URL
  \url{http://papers.nips.cc/paper/3423-bounding-performance-loss-in-approximate-mdp-homomorphisms.pdf}.

\bibitem[Towell and Shavlik(1993)]{towell1993extracting}
Geoffrey~G Towell and Jude~W Shavlik.
\newblock Extracting refined rules from knowledge-based neural networks.
\newblock \emph{Machine learning}, 13\penalty0 (1):\penalty0 71--101, 1993.

\bibitem[Tschiatschek et~al.(2018)Tschiatschek, Arulkumaran, St{\"{u}}hmer, and
  Hofmann]{hofmann2018pomdp}
Sebastian Tschiatschek, Kai Arulkumaran, Jan St{\"{u}}hmer, and Katja Hofmann.
\newblock Variational inference for data-efficient model learning in {POMDPs}.
\newblock \emph{CoRR}, abs/1805.09281, 2018.

\bibitem[van~den Oord et~al.(2017)van~den Oord, Vinyals, and
  kavukcuoglu]{vandenoord2017vqvae}
Aaron van~den Oord, Oriol Vinyals, and koray kavukcuoglu.
\newblock Neural discrete representation learning.
\newblock In I.~Guyon, U.~V. Luxburg, S.~Bengio, H.~Wallach, R.~Fergus,
  S.~Vishwanathan, and R.~Garnett, editors, \emph{Advances in Neural
  Information Processing Systems 30}, pages 6306--6315. Curran Associates,
  Inc., 2017.

\bibitem[Veness et~al.(2011)Veness, Ng, Hutter, Uther, and
  Silver]{veness2011aixi}
Joel Veness, Kee~Siong Ng, Marcus Hutter, William Uther, and David Silver.
\newblock A monte-carlo aixi approximation.
\newblock \emph{J. Artif. Int. Res.}, 40\penalty0 (1):\penalty0 95–142,
  January 2011.
\newblock ISSN 1076-9757.

\bibitem[Wang et~al.(2017)Wang, Zhang, Ororbia, Alexander, Xing, Liu, and
  Giles]{wang2017empirical}
Qinglong Wang, Kaixuan Zhang, II~Ororbia, G~Alexander, Xinyu Xing, Xue Liu, and
  C~Lee Giles.
\newblock An empirical evaluation of recurrent neural network rule extraction.
\newblock \emph{arXiv preprint arXiv:1709.10380}, 2017.

\bibitem[Wang et~al.(2018)Wang, Zhang, Ororbia, Alexander, Xing, Liu, and
  Giles]{wang2018comparison}
Qinglong Wang, Kaixuan Zhang, II~Ororbia, G~Alexander, Xinyu Xing, Xue Liu, and
  C~Lee Giles.
\newblock A comparison of rule extraction for different recurrent neural
  network models and grammatical complexity.
\newblock \emph{arXiv preprint arXiv:1801.05420}, 2018.

\bibitem[Weiss et~al.(2017)Weiss, Goldberg, and Yahav]{weiss2017extracting}
Gail Weiss, Yoav Goldberg, and Eran Yahav.
\newblock Extracting automata from recurrent neural networks using queries and
  counterexamples.
\newblock \emph{arXiv preprint arXiv:1711.09576}, 2017.

\bibitem[Yang et~al.(2020)Yang, Zhang, Morcos, Pineau, Abbeel, and
  Calandra]{yang2020plan2vec}
Ge~Yang, Amy Zhang, Ari~S. Morcos, Joelle Pineau, Pieter Abbeel, and Roberto
  Calandra.
\newblock Plan2vec: Unsupervised representation learning by latent plans.
\newblock In \emph{Proceedings of The 2nd Annual Conference on Learning for
  Dynamics and Control}, volume 120 of \emph{Proceedings of Machine Learning
  Research}, pages 1--12, 2020.
\newblock arXiv:2005.03648.

\bibitem[Zhang et~al.(2018)Zhang, Lerer, Sukhbaatar, Fergus, and
  Szlam]{zhang2018composable}
Amy Zhang, Adam Lerer, Sainbayar Sukhbaatar, Rob Fergus, and Arthur Szlam.
\newblock Composable planning with attributes.
\newblock In \emph{Proceedings of the 35th International Conference on Machine
  Learning, {ICML} 2018, Stockholmsm{\"{a}}ssan, Stockholm, Sweden, July 10-15,
  2018}, volume~80, pages 5837--5846. JMLR.org, 2018.

\bibitem[Zhang et~al.(2020{\natexlab{a}})Zhang, Lyle, Sodhani, Filos,
  Kwiatkowska, Pineau, Gal, and Precup]{zhang2020invariant}
Amy Zhang, Clare Lyle, Shagun Sodhani, Angelos Filos, Marta Kwiatkowska, Joelle
  Pineau, Yarin Gal, and Doina Precup.
\newblock Invariant causal prediction for block {MDPs}.
\newblock In \emph{International Conference on Machine Learning (ICML)},
  2020{\natexlab{a}}.

\bibitem[Zhang et~al.(2020{\natexlab{b}})Zhang, McAllister, Calandra, Gal, and
  Levine]{zhang2020dbc}
Amy Zhang, Rowan McAllister, Roberto Calandra, Yarin Gal, and Sergey Levine.
\newblock Learning invariant representations for reinforcement learning without
  reconstruction.
\newblock \emph{arXiv preprint arXiv:2006.10742}, 2020{\natexlab{b}}.

\end{thebibliography}

\end{document}